\documentclass{article}

\usepackage{PRIMEarxiv}

\usepackage[utf8]{inputenc} % allow utf-8 input
\usepackage[T1]{fontenc}    % use 8-bit T1 fonts
\usepackage{hyperref}       % hyperlinks
\usepackage{url}            % simple URL typesetting
\usepackage{booktabs}       % professional-quality tables
\usepackage{amsfonts}       % blackboard math symbols
\usepackage{nicefrac}       % compact symbols for 1/2, etc.
\usepackage{microtype}      % microtypography
\usepackage{lipsum}
\usepackage{fancyhdr}       % header
\usepackage{graphicx}       % graphics
\graphicspath{{media/}}     % organize your images and other figures under media/ folder

\usepackage{tikz}
\usetikzlibrary{arrows,automata,positioning}

\usepackage{mathtools}
\usepackage{xspace}

\usepackage{hyperref}

\usepackage[utf8]{inputenc} % allow utf-8 input
\usepackage[T1]{fontenc}    % use 8-bit T1 fonts
\usepackage{hyperref}       % hyperlinks
\usepackage{url}            % simple URL typesetting
\usepackage{booktabs}       % professional-quality tables
\usepackage{amsfonts}       % blackboard math symbols
\usepackage{nicefrac}       % compact symbols for 1/2, etc.
\usepackage{microtype}      % microtypography
\usepackage{xcolor}         % colors
\usepackage{answers}

\usepackage[ruled,vlined]{algorithm2e}

\title{Chebyshev Particles
}

\author{
  Xiongming Dai  \\
  Division of Computer Science and Engineering \\
  Louisiana State University \\
  Baton Rouge,LA70803, USA\\
  \texttt{\{xdai2\}@email} \\
   \And
  Gerald Baumgartner \\
  Division of Computer Science and Engineering \\
  Louisiana State University \\
  Baton Rouge,LA70803, USA\\
  \texttt{\{gb\}@email} \\
}

\begin{document}
\maketitle

\begin{abstract}
 Markov chain Monte Carlo (MCMC) provides a feasible method for inferring Hidden Markov models, however, it is often computationally prohibitive, especially constrained by the curse of dimensionality, as the Monte Carlo sampler traverses randomly taking small steps within uncertain regions in the parameter space. %Deterministic approximation is a popular mechanism to find an asymptotically optimal region of interest in space-filling designs of computer experiments, where deterministic points distributed geometrically reflect the special representations of the region. 
We are the first to consider the posterior distribution of the objective as a mapping of samples in an infinite-dimensional Euclidean space where deterministic submanifolds are embedded and propose a new criterion by maximizing the weighted Riesz polarization quantity, to discretize rectifiable submanifolds via pairwise interaction.  We study the characteristics of Chebyshev particles and embed them into sequential MCMC, a novel sampler with a high acceptance ratio that proposes only a few evaluations. We have achieved high performance from the experiments for parameter inference in a linear Gaussian state-space model with synthetic data and a non-linear stochastic volatility model with real-world data.

\end{abstract}

% keywords can be removed

\section{Introduction}
%\section{Introduction}
%2003-an efficient algorithm for constructing optimal desing of computer experiment

%2006—Book_Design and Modeling for Computer Experiments.... for second paragraph.
%2010-Generalized Latin Hypercube Design for computer experiment.
%2015-sequential exploration of complex surfaces using minimum energy desings.
%2015-space-filling desings for computer experiments
%2016-GU_dissertation_2016 Minimum Energy designs_extensions
%2017-deterministic sampling of expensive posteriors

%uniquely should be added!
Markov chain Monte Carlo methods \cite{metropolis1953equation,hastings1970monte} allow researchers to replace the unobserved latent variables with simulated variables for Bayesian analysis \cite{geman1984stochastic,besag1991bayesian}. It relieves the burden of evaluating the likelihood function unconditionally on the unobserved latent variables to allow a focus on the conditional likelihood function \cite{smith1993bayesian,doucet2001introduction,smith2013sequential}. However, although the sample is drawn after a ``burn-in" period, it is still uncertain whether the sample is the output at convergence \cite{plummer2006coda}. Using the output of an MCMC algorithm that has not converged may lead to incorrect inferences on the target distribution at hand. In addition, the idea behind a Monte Carlo sampler is to randomly "walk around" in the parameter space, which may lead to a low acceptance rate of samples and generate duplicates, especially for a high dimensional space~\cite{plummer2006coda}. Thus, an adequate amount of samples are obtained at the expense of a large computational effort.

Points on a design space that minimize certain energy functions often have desirable properties such as good separation and adequate covering and sparsely reflect special representations of the space \cite{damelin2005point,hardin2012quasi,hardin2005minimal,borodachov2008asymptotics,borodachov2014low}. Here, we consider the posterior distribution of the objective function as a mapping of samples in an infinite-dimensional Euclidean space, where deterministic submanifolds are embedded, and propose a new criterion by maximizing weighted Riesz polarization quantity to discretize rectifiable submanifolds via particle interaction. This gives rise to equilibrium points that are useful for a variety of applications, especially for high-dimensional sampling. We study the characteristics of deterministic points, termed Chebyshev particles, and embed them into the MCMC\@. We propose a new sampler with few evaluations and a high acceptance ratio. In our experiments, we have achieved high performance for parameter inference in a linear Gaussian state-space model with synthetic data as well as a non-linear stochastic volatility model with real-world data.

In this paper, we concentrate on the analysis of our new criterion, and on how to improve the acceptance ratio of MCMC with fewer evaluations of constraints. We present an efficient algorithm to deterministically sample the target distribution by maximizing the weighted Riesz polarization quantity, from which the particles sparsely represent rectifiable geometrical manifolds with few samplings for approximating the objective posterior distribution.  

In Section 2, we present a brief introduction to the discrete energy on rectifiable sets. Here, we develop our new criterion by maximizing the weighted Riesz polarization quantity and focus on the bounds and asymptotic behavior, and covering radius. For Section 3, we propose a novel sampler, where Chebyshev particles are embedded and the discretized deterministic submanifolds inherit the special representations of the sampling space. Then, we present the pipeline for sequentially sampling the Chebyshev particles and for embedding them into the particle Metropolis-Hastings algorithm for hidden Markov models. In Section 4, we validate the algorithm with practical experiments and present their performance and error analysis. The summary of our contributions is outlined in Section 5.
\section{Weighted Riesz Polarization Criterion}

In this section, we provide the main idea of the discrete energy on rectifiable high-dimensional manifolds and propose a new criterion by maximizing the weighted Riesz polarization quantity. We then study the asymptotic behavior of the corresponding configuration, the bounds and the covering radius for this quantity. %All these special features will be inherited when we embed weighted Riesz particles into sequential MCMC, the new sampler shows high performance within few samples, which is verified by experiment for hidden Markov models in Section 4.

\subsection{Discrete Weighted Riesz Polarization}
Let $\Omega$ denote a compact set in $\mathbb{R}^d$ whose $d$-dimensional Borel measure, $\mathbb{B}_d(\Omega )\subset (\Omega,\mathbb{R}^d)$, is finite, and $K$ denote a bi-Lipschitz mapping from $\Omega \times \Omega $ to $\mathbb{R}^d$, for a collection of $n(\geq 2)$ distinct points of configuration in $\Omega$, let $X_{1:n}={x_1,...,x_n}$, we define the energy of $X_{1:n}$ to be
\begin{equation}{\label{b2}}
E(X_{1:n}):=\sum_{i=1}^{n}\sum_{j=1,j\neq i}^{n}K(x_i,x_j)=\sum_{i\neq j}^{}K(x_i,x_j),
\end{equation}
and let 
\begin{equation}{\label{b3}}
\mathcal{E}(\Omega,n):=\text{inf}\{E(X_{1:n}):X_{1:n}\subset \Omega,\left | X_{1:n} \right |=n  \}
\end{equation}
be the minimal discrete $n$-point energy of the configuration in $\Omega$, where $\left | X_{1:n} \right |$ represents the cardinality of the set $X_{1:n}$. (\uppercase\expandafter{\romannumeral1}) For $K(x_i,x_j)=-\text{log}\parallel x_i-x_j\parallel $, it was first proposed by M.Fekete who explored the connection between polynomial interpolation and discretized manifolds \cite{fekete1923verteilung}. In computational complexity theory, % { \color{black}Smale} 
Smale \cite{smale1998mathematical} proposed the $7th$ problem in his list of "Mathematical problems for the next century" that how to design a polynomial time algorithm for generating ``nearly" optimal logarithmic energy points $X_{1:n}^{*}$, also called Fekete points, on the unit sphere in $\mathbb{R}^3$ that satisfy $E(X_{1:n}^{*})-\mathcal{E}(\mathbb{S}^2,n)\leq \footnote{\text{In this paper} { {$C_i$}} $\in \mathbb{R}^{+},i=1,2,...m $\text{ denote different constants.}}{ {C_1}} \cdot \text{log}n$ for some universal constant ${ \color{black}C_1}$; (\uppercase\expandafter{\romannumeral2}) when $K(x_i,x_j)=\frac{1}{\parallel x_i-x_j\parallel^m}, m\in \mathbb{R}^{+}$, let $\mathcal{E}_{m}(\Omega ,n)$ denote the Riez $m$-energy, by Taylor's formula, for any $m\in (0,+\infty )$, we have
\begin{equation}{\label{bb}}
\begin{split}
\lim_{m\rightarrow 0^+}\mathcal{E}_{m}(\Omega,n)  =\lim_{m\rightarrow 0^+}\frac{n(n-1)+m\mathcal{E}_\text{log}(\Omega ,n)+\mathcal{O}(m)}{m} =\mathcal{E}_{\log}(\Omega,n).
\end{split}
\end{equation}
Consequently, the Fekete points set $X_{1:n}^{(m)}$ can be considered as limiting cases of point sets that minimize the discrete Riesz energy, which is widely used to discretize manifolds via particle interactions in Euclidean space \cite{borodachov2008asymptotics,hardin2004discretizing}.

From the perspective of statistical high-dimensional sampling, we consider $d$ sufficiently large and propose the maximum  weighted Riesz polarization criterion with
\begin{align}
\label{eqn:eqlabel10}
\begin{split}
 & \mathcal{E}_{\beta} (\Omega)=\max_{\Omega} \min_{x_i,x_j}\left \{ \sum_{i=1}^{n-1}\sum_{j=i+1}^{n}\frac{\omega(x_i,x_j)}{\parallel x_i-x_j\parallel^m}  \right \}^{\frac{1}{m}}, \omega (x_i,x_j)\propto e^{\left [ \alpha \cdot \gamma (x_i) \gamma (x_j) +\beta \cdot \parallel x_i-x_j\parallel \right ]^{-\frac{m}{2d}}}.\\
 % & \omega (x_i,x_j)=\left [ \gamma (x_i) \gamma (x_j) +\beta \parallel x_i-x_j\parallel \right ]^{-\frac{m}{2d}}\omega' (x_i,x_j). \\
 %&  \omega' (x_i,x_j)=\Phi (\frac{\parallel x_i-x_j\parallel}{r_N}), \\
 %& r_N=C_NN^\frac{-1}{d}, \exists \beta, C_N\in R^+. \\
\end{split}
\end{align}
As $m\rightarrow \infty$, the formulation is convex under mild conditions, the denominator approximates $\parallel x_i-x_j\parallel$, thus, our criterion inherits the properties of Riesz energy, termed as weighted Riesz polarization criterion.
To obtain a finite collection of point sets that are distributed according to a specified non-uniform density such as might be used as points for weighted integration or design of complex surfaces where more points are required in regions with higher curvature, we introduce $\omega (x_i,x_j)$ in ~\eqref{eqn:eqlabel10}, where $\gamma (x)\propto -\ln{f(x)}$, $\parallel x_i-x_j\parallel$ is included to ensure that is locally bounded for $\alpha=-1$, $\beta$ is the local discrepancy coefficient and is positive to balance off the local conflict with the distributed points when short-range interactions between points are the dominator. Thus, given a proper distribution $f(x)$, we can use $\mathcal{E}_{\beta} (\Omega)$ to generate a sequence of $n$-point configurations that are "well-separated" and have asymptotic distribution $f(x)$.
 
 Our weighted Riesz polarization $\mathcal{E}_{\beta} (\Omega, N)$ is continuous and derivable with respect to the parameter $\beta \subset \mathbb{R}$ from ~\eqref{eqn:eqlabel10}, it provides a more flexible and versatile framework when we discretize the submanifolds via particle interactions.
%The corresponding energy criterion from ~\eqref{eqn:eqlabel10} is denoted by $\varepsilon_{\beta_1} (\Omega ,N)$.
%{ \color{black}We will present some special features of our model for the following part.}
 %1997-Article_DistribuingMany points on a Sphere
\subsection{Asymptotics for Extremal Weighted Riesz Polarization Criterion}
{\textbf {Properties of $\omega(x_i,x_j)$ }}: (\uppercase\expandafter{\romannumeral1}) $\omega(x_i,x_j)$ is continuous as a function of $\gamma (x)\propto -\ln{f(x)}$ when $\beta \leq \beta_0$; it is a positive constant when $\beta \geq \beta_1$; (\uppercase\expandafter{\romannumeral2}) There exists a neighborhood set $P'$, where $x_i',x_j'  \in P',  \omega(x_i',x_j')$ is bounded and larger than zero; (\uppercase\expandafter{\romannumeral3}) $\omega(x_i,x_j)$ is bounded on any closed and compact metric space $\Omega$.

Assume the compact set $\Omega \subset \mathbb{R}^d$, for high dimension $m>d$, we define the generalized Borel measure on sets $\mathbb{S} \subset \Omega$ with $\mathcal{U}_d^m(\mathbb{S}) :=  \int _\mathbb{S} \omega(x_i,x_j)d\mathcal{U}_d(x).$
\iffalse
\begin{equation*}
\mathcal{U}_d^m(\mathbb{S}) :=  \int _\mathbb{S} \omega(x_i,x_j)d\mathcal{U}_d(x).
\end{equation*}
\fi
It is bounded and the corresponding normalized form: $u_d^m(\mathbb{S}):=  \mathcal{U}_d^m(\mathbb{S})/ \mathcal{U}_d^m({\Omega}).
$
\iffalse
\begin{equation*}
u_d^m(\mathbb{S}):=  \mathcal{U}_d^m(\mathbb{S})/ \mathcal{U}_d^m({\Omega}).
\end{equation*}
\fi

$\textbf{Measure Metric}$ Consider a high dimensional space $\mathbb{R}^d, d \ge 2$, for $m>d$, let $\mu(\sigma \text{-algebra}):=  \cup_{d=2}^{\infty }\{ \parallel x_i-x_j\parallel^d\}$, represent a Borel measure from the $\sigma $-algebra on $\Omega$, a measure $\phi  $ in $\Omega_i$ is a non-negative $\sigma $-algebra set function defined on $\mu(\sigma \text{-algebra})$ and finite on all compact sets $\Omega_i \subset \Omega,i\in [1,n]$. If $\phi  < \infty $, then the measure $\phi $ is called finite. Generally, for the smallest $\sigma $-algebra, containing all compact subsets of $\Omega_i$. 

We have the following novel version of the Poppy-Seed Bagel Theorem \cite{borodachov2019discrete} for the maximum weighted Riesz polarization using the measure theoretics \cite{hunter2011measure}. 

%%%%%%%%for Riesz particles%%%%%%%%%%%%%
% chapter11 page 479 is the main theory for the old version:
%chapter 8, Page393, the proof of the classical!
%Theorem 11.1.2 Page 481
%%%%%%%for Chebyshev Particles%%%%%%%%%%
%Theorem 14.9.2 Page 588 book
{\textbf {Theorem 2.2.1.}} Given a distribution $f(x)$ with respect to a compact and $d$-rectifiable set $\Omega$ embedded in Euclidean space $\mathbb{R}^d$, $\omega(x_i,x_j)>0$ is bounded and continuous on the closed Borel sets $\mathbb{S} \subset \Omega \times \Omega $, for $m>d$, the configuration on $\Omega$ from $\mathcal{E}_{\beta}(\Omega)$ where the $N$-point interacts via the $K_{\beta}(x_i,x_j)$ potential, have
\iffalse
For a given density function $f(x)$, draw a set of well-separated samples of $N$-point configurations on a compact space $\Omega \subset \mathbb{R}^d $ that is distributed in the limit (as $N\rightarrow \infty$) with the density $f(x)$, and there exists a set $Q \subset  \mathbb{R}$, where $\beta \in \mathbb{R} $. If $m>d$, and $\omega(x_i,x_j)$ is a bounded, continuous and positive function on the closed Borel sets $\mathbb{S} \subset \Omega \times \Omega $, then
\begin{equation}\label{eqn:eqlabel13}
\lim_{N\rightarrow \infty }\frac{\varepsilon_{\beta} (\Omega ,N)}{N^{1+m^2/d^2}}=\frac{C_{m^2,d^2}}{[\mathcal{H}_d^m(\mathbb{S})]^{m^2/d^2}}
\end{equation}
\begin{equation}\label{eqn:eqlabel13}
\begin{cases}
 &\lim_{N\rightarrow \infty }\lim_{\beta^{-} \rightarrow \beta_0 }\frac{\varepsilon_{\beta} (\Omega ,N)}{N^{\frac{1}{m}+\frac{1}{d}}}=\frac{{ \color{black}C_2}}{[\mathcal{U}_d^m(\mathbb{S})]^{\frac{1}{d}}}, \\ 
 &\lim_{N\rightarrow \infty }\lim_{\beta^{+} \rightarrow \beta_1 }\frac{\varepsilon_{\beta} (\Omega ,N)}{N^{\frac{2}{m}+\frac{2}{d}}}=\frac{{ {C_3}}}{[\mathcal{U}_d^m(\mathbb{S})]^{\frac{2}{d}}}. 
 \end{cases}
\end{equation}
\fi
\begin{equation}\label{eqn:eqlabel13}
 \lim_{n\rightarrow \infty }\frac{\mathcal{E}_{\beta} (\Omega)}{n^{\frac{1}{m}+\frac{1}{d}}}=\frac{{ \color{black}C_2}}{[\mathcal{U}_d^m(\mathbb{S})]^{\frac{1}{d}}}.
 \end{equation}
Moreover, if $\mathcal{U}_d^m(\mathbb{S})>0$, any configuration $X_{1:n},n>1$ generated by asymptotically maximizing the weighted Riesz polarization is uniformly distributed with respect to $\mathcal{U}_d$, that is, 
\begin{equation}\label{eqn:eqlabel14}
\lim_{n\rightarrow \infty }\frac{1}{n}\sum_{i=1,i\neq j}^{n }\parallel x_i-x_j\parallel =u_d^m(\mathbb{S}).
\end{equation}
\iffalse
{\textbf {Theorem 2 }} For the compact space $\Omega$ subsets in a two-dimensional $C^1$-manifold, where $\mathcal{H}_2(\Omega)=0$, then

\begin{equation}\label{eqn:eqlabel15}
\lim_{N\rightarrow \infty }(\lim_{m\rightarrow d}\frac{\varepsilon (\Omega ,N)}{N^2{\text{log}}N})=\triangle V
\end{equation}

Similarly, if $\mathcal{H}_2(\Omega)>0$, any configuration $\omega (x_1,x_2,...x_N),N>1$ generated from the asymptotically minimize the criterion $\varepsilon (\Omega ,N)$ is uniformly distributed with respect to $\mathcal{H}_d$, that is, ~\eqref{eqn:eqlabel14} still holds for $m=d$.
\fi

$\textbf{Proof of Theorem 2.2.1}$ We divide the proof of Theorem 2.2.1 into two parts; The proof works by induction with Lemma 2.2.2 for  %~\eqref{eqn:eqlabel13},
(4) in the main text, and Lemmas 2.2.3, 2.2.4, 2.2.5, 2.2.6 and 2.2.7 for (5) in the main text. The second part will introduce the subadditivity and superadditivity properties using the measure theoretics \cite{hunter2011measure}.
%~\eqref{eqn:eqlabel14}.
%We first use limit and series theory to prove ~\eqref{eqn:eqlabel13} by Lemma 2.2.2, then consider the problem of asymptotically minimizing weighted Riesz energy in a metric space to prove ~\eqref{eqn:eqlabel14} by the extension of derivation from Lemma 2.2.3 to Lemma 2.2.7. 

$\textbf{Lemma 2.2.2.}$  Given a distribution $f(x)$ with respect to $d$-rectifiable set $\Omega$ embedded in Euclidean space, $\omega(x_i,x_j)>0$ is bounded and continuous on the closed Borel sets $\mathbb{S} \subset \Omega \times \Omega $, for $m>d$ and $\beta \in(-\infty ,\beta_0]\cup [\beta_1,+\infty ),\beta_0,\beta_1,{ \color{black}C_3} \in \mathbb{R}$, the maximal weighted Riesz polarization configuration on $\Omega$ from $\mathcal{E}_{\beta}(\Omega)$ where the $n$-point interacts via the $K_{\beta}(x_i,x_j)$ potential, have
\begin{equation}\label{eqn:eqlabel13dd}
\lim_{\substack{{n} \to \infty}}\lim_{\beta^{-} \to \beta_0} \frac{\mathcal{E}_{\beta} (\Omega)}{n^{\frac{1}{m}+\frac{1}{d}}}=\frac{{ \color{black}C_3}}{[\mathcal{U}_d^m(\mathbb{S})]^{\frac{1}{d}}},\ \ \lim_{\substack{{n} \to \infty}}\lim_{\beta^{+} \to \beta_1} \frac{\mathcal{E}_{\beta} (\Omega)}{n^{\frac{1}{m}+\frac{1}{d}}}=\frac{{ \color{black}C_3}}{[\mathcal{U}_d^m(\mathbb{S})]^{\frac{1}{d}}}. 
\end{equation}
$\textbf{Proof}$ $\mathcal{E}_{\beta}(\Omega)$ is strictly decreasing as $\beta$ increases, this monotonicity makes it possible to analyze the asymptotics and extend it into high-dimensional sampling on the compact space $\Omega$ under mild assumptions. Let $h(\beta'):=\lim_{n\rightarrow \infty }\lim_{\beta \rightarrow \beta' }\mathcal{E}_{\beta} (\Omega)$, 
 \begin{equation*}{\label{h12}}
h(\beta)=\sum_{i=1}^{n}\sum_{j=1,j\neq i}^{n}\left[K_{\beta}(x_i,x_j)\right ]^{\frac{1}{m}}.
\end{equation*}
$K_{\beta}(x_i,x_j)$ is also strictly decreasing as $\beta$ increases, we firstly focus on $\beta \in(-\infty ,\beta_0]\cup [\beta_1,+\infty ),\beta_0,\beta_1 \in \mathbb{R}$ , then relax this assumption later, define
\begin{equation*}\label{eqn:eqlabel18}
K_{\beta'}(x_i,x_j):=\lim_{\beta \rightarrow \beta' }\frac{ \omega(x_i,x_j)}{\parallel x_i-x_j\parallel^m},\omega(x_i,x_j)> 0,
\end{equation*}
if $\beta \leq \beta_0$ is sufficiently small such that
\begin{equation}\label{eqn:eqlabel19a}
\gamma (x_i)\gamma (x_j)\gg\beta_0 \cdot \parallel x_i-x_j\parallel,
\end{equation}
then,
\begin{equation}\label{eqn:eqlabel19}
K_{\beta_0^{-}}(x_i,x_j):=\lim_{\beta\rightarrow \beta_0^{-}}K_{\beta}(x_i,x_j)=\frac{e^{\left [- \gamma (x_i) \gamma (x_j) \right ]^{-\frac{m}{2d}}}}{\parallel x_i-x_j\parallel^m}.
\end{equation}
From Taylor's theorem
\begin{equation*}\label{eqn:eqlabel19v}
e^z=1+z+\frac{z^2}{2!}+\cdots +\frac{z^{k'}}{k'!},k'\to \infty, z\in \mathbb{R}.
\end{equation*}
Let $z={\left [- \gamma (x_i) \gamma (x_j) \right ]^{-\frac{m}{2d}}}$, substitute ~\eqref{eqn:eqlabel19a} into ~\eqref{eqn:eqlabel19}, 
\begin{equation*}
\label{eqn:eqlabel20d2a}
\begin{split}
K_{\beta_0^{-}}(x_i,x_j)&=\frac{1+z+\frac{z^2}{2!}+\cdots +\frac{z^{k'}}{k'!}}{\parallel x_i-x_j\parallel^m} \ge \frac{1}{\parallel x_i-x_j\parallel^m}+ \frac{\left[-\beta_0 \cdot \parallel x_i-x_j\parallel\right]^{-\frac{m}{2d}}}{\parallel x_i-x_j\parallel^m}+...\\
&+\frac{\left[-\beta_0 \cdot \parallel x_i-x_j\parallel\right]^{\frac{-{m{k'}}}{2d}}}{{k'}!\parallel x_i-x_j\parallel^m}.
\end{split}
\end{equation*}
For $m>d$, the right-hand side terms are belonging to the classical Riesz-kernel model, from  the Poppy-Seed Bagel Theorem \cite{borodachov2019discrete}, there exists a ${{ \color{black}C_4}}$,
\begin{equation*}\label{eqn:eqlabeld19v}
K_{\beta_0^{-}}(x_i,x_j)=\frac{{ \color{black}C_4}}{[\mathcal{U}_d^m(\mathbb{S})]^{\frac{m}{d}}} \cdot n^{1+\frac{m}{d}}.
\end{equation*}
Thus,
\begin{equation*}\label{eqn:eqlabeld19dddv}
h(\beta_0^{-}):=\sum_{i=1}^{n}\sum_{j=1,j\neq i}^{n}\left[K_{\beta_0^{-}}(x_i,x_j)\right ]^{\frac{1}{m}}=\frac{{ \color{black}C_4}}{[\mathcal{U}_d^m(\mathbb{S})]^{\frac{1}{d}}}\cdot n^{\frac{1}{m}+\frac{1}{d}}.
\end{equation*}
%The corresponding energy criterion from ~\eqref{eqn:eqlabel10} is denoted by $\varepsilon_{\beta_0} (\Omega ,N)$.
Similarly, if $\beta \geq \beta_1$ is  sufficiently large such that $ \gamma (x_i) \gamma (x_j) \ll \beta_1 \cdot \parallel x_i-x_j\parallel$, then
\begin{align}
\label{eqn:eqlabel20ew}
\begin{split}
   K_{\beta_1^{+}}(x_i,x_j) :&=\lim_{\beta\rightarrow \beta_1^{+}}K_{\beta}(x_i,x_j)=\frac{e^{\left [ \beta_1 \parallel x_i-x_j\parallel \right ]^{-\frac{m}{2d}}}}{\parallel x_i-x_j\parallel^m}= \frac{1}{\parallel x_i-x_j\parallel^m}+\frac{\left[\beta_1 \cdot \parallel x_i-x_j\parallel\right]^{-\frac{m}{2d}}}{\parallel x_i-x_j\parallel^m}\\
   &+...+ \frac{\left[\beta_1 \cdot \parallel x_i-x_j\parallel\right]^{\frac{-{m{k'}}}{2d}}}{{k'}!\parallel x_i-x_j\parallel^m}.
\end{split}
\end{align}
It provides a flexible framework to prove the asymptotics of the proposed weighted Riesz polarization criterion for ~\eqref{eqn:eqlabel20ew} that we will frequently use for the following lemma and related proof.

 For $m>d$, the right-hand side terms belong to the classical Riesz-kernel model, from  the Poppy-Seed Bagel Theorem \cite{borodachov2019discrete}, there exists a ${{ \color{black}C_5}}$,
\begin{equation*}\label{eqn:eqlabeld1dd9v}
K_{\beta_1^{+}}(x_i,x_j)=\frac{{ \color{black}C_5}}{[\mathcal{U}_d^m(\mathbb{S})]^{\frac{m}{d}}} \cdot n^{1+\frac{m}{d}}.
\end{equation*}
Thus,
\begin{align}
\label{eqn:eqlabeld19dvd}
\begin{split}
   h(\beta_1^{+})&:=\sum_{i=1}^{n}\sum_{j=1,j\neq i}^{n}\left[K_{\beta_1^{+}}(x_i,x_j)\right ]^{\frac{1}{m}}=\frac{{ \color{black}C_5}}{[\mathcal{U}_d^m(\mathbb{S})]^{\frac{1}{d}}}\cdot n^{\frac{1}{m}+\frac{1}{d}}.
\end{split}
\end{align}

As $h({\beta})$ is strictly decreasing, and continuous and derivative for $\beta\in \mathbb{R}$, Consequently, There exists a ${{ \color{black}C_3}}$, 
\begin{equation*}\label{eqn:eqlabel1dx3}
 \lim_{n\rightarrow \infty }\frac{\mathcal{E}_{\beta} (\Omega)}{n^{\frac{1}{m}+\frac{1}{d}}}=\frac{{ \color{black}C_3}}{[\mathcal{U}_d^m(\mathbb{S})]^{\frac{1}{d}}}.
 \end{equation*}
Thus, ~\eqref{eqn:eqlabel13dd} holds.

From Lemma 2.2.2, as $n \to \infty$, the approximation of $\mathcal{E}_{\beta} (\Omega)$ is not correlated with $\beta$.
That is, we are assuming that $\beta$ approximates a specific real value, and for the convenience of introducing Taylor's theorem to derive, it does not affect the final limit value of $\mathcal{E}_{\beta} (\Omega)$ for $n \to \infty$.
  
Analogous to the proof of classical Poppy-Seed Bagel Theorem \cite{borodachov2019discrete}, we define 
\begin{equation*}\label{eqn:eqlabel17a}
\mathcal{M}(d):= 1+\frac{m}{d},n \geq 2.
\end{equation*}
$\lambda (n):=n^{\mathcal{M}(d)}$, for $n\geq 2$, $\lambda (1):=1$.
And define
\begin{equation}\label{eqn:eqlabel17}
\psi_{m,d} (\Omega):=\lim_{n\rightarrow \infty }\frac{\mathcal{E}^m_\beta (\Omega)}{\lambda (n)},
\end{equation}
let $\psi_{m,d}^{\text{inf}} (\Omega)={\text{inf}}(\psi_{m,d} (\Omega))$, $\psi_{m,d}^{\text{sup}} (\Omega)={\text{sup}}(\psi_{m,d} (\Omega))$ and decompose the $d$-rectifiable set $\Omega$ into different subsets $\Omega_i, i \in \mathbb{R}^+$, satisfying $\cup_{i=1}^\infty {\Omega_i}=\Omega$. %{ \color{black}Before we derive this limit, it is necessary to expand the Lemma 8.6.3 (page 387 \cite{borodachov2019discrete}) to a general case, and this is very useful for the later proofs.} 
%chapter 8.5 poppy seed bagel theorems: page 379 
%the proof is (chapter 8.5 page 382- \cite{borodachov2019discrete})  
%the true proof begins from chapter 8.7 page 393-398

{\textbf{Lemma 2.2.3.}} \cite{borodachov2019discrete} $\exists \alpha_1, \alpha_2 \in \mathbb{R}^+$,  $\mathcal{M}(d)$ is continuous and derivative for $d \in \mathbb{R}^+$, the function $U(t)=\min\{\alpha_1t^{\mathcal{M}(d)-1},\alpha_2(1-t)^{\mathcal{M}(d)-1}\}$
\iffalse
\begin{equation*}\label{eqn:eqlabel21}
U(t)=\alpha_1t^{\mathcal{M}(d)-1}+\alpha_2(1-t)^{\mathcal{M}(d)-1}
\end{equation*}
\fi
has the maximum for $t \in [0,1]$ where occurs at the points $t^*:=\frac{1}{1+(\frac{\alpha_1}{\alpha_2})^\frac{1}{\mathcal{M}(d)-1}}$ with $U(t^*)= \left[ \alpha_2^{\frac{-1}{1-\mathcal{M}(d)}}+\alpha_1^{\frac{-1}{1-\mathcal{M}(d)}}\right]^{1-\mathcal{M}(d)}$.
\iffalse
\begin{equation*}\label{eqn:eqlabel22}
t^*:=\frac{1}{1+(\frac{\alpha_1}{\alpha_2})^\frac{1}{\mathcal{M}(d)-2}}
\end{equation*}
with
\begin{equation*}\label{eqn:eqlabel2233}
U(t^*)= \left[ \alpha_2^{\frac{-1}{\mathcal{M}(d)-2}}+\alpha_1^{\frac{-1}{\mathcal{M}(d)-2}}\right]^{2-\mathcal{M}(d)}.
\end{equation*}
\\
\fi
\\
The proof is straightforward from the first order derivative of the function $\frac{dU(t)}{dt}=0$ \cite{borodachov2019discrete}. 

We will introduce the subadditivity and superadditivity properties as follows.

%(page 382 Lemma8.6.4) and page 394 Lemma8.7.3 
%Lemamma:11.1.5 page 482 
  {\textbf{Lemma 2.2.4.}} $\exists \Omega_j,\Omega_k \subset \Omega$, and $\Omega_j,\Omega_k \not\subset  \emptyset $, $j\neq k$, $\mathcal{M}(d) > 1$ is continuous and derivative for  $d \in \mathbb{R}^+$, let $\alpha_3=\frac{1}{1-\mathcal{M}(d)}$ for $m >d$,
\begin{equation}\label{eqn:eqlabel27dd}
\psi_{m,d}^{\text{inf}} (\Omega_j \cup \Omega_k)^{\alpha_3} \le \psi_{m,d}^{\text{inf}} (\Omega_j)^{\alpha_3}+\psi_{m,d}^{\text{inf}} (\Omega_k)^{\alpha_3}.
\end{equation}
{\textbf{Proof}}
If $\psi_{m,d}^{\text{inf}}(\Omega_j)$ or $\psi_{m,d}^{\text{inf}} (\Omega_k)$ equals zero, or one of the quantities $\psi_{m,d}^{\text{inf}}(\Omega_j)$ or $\psi_{m,d}^{\text{inf}} (\Omega_k)$ approximates infinite, as the size of set increase, $\mathcal{E}_{\beta}^m (\Omega)$ will increase, the lemma holds.

Hereafter we follow an argument in \cite{borodachov2019discrete}, we consider the general case of $\psi_{m,d}^{\text{sup}} (\Omega_j)\in (0,\infty ),\psi_{m,d}^{\text{sup}} (\Omega_k)\in (0,\infty )$, the distance of two set is defined with $r:=\left \| a_i-b_j \right \|,a_i \in \Omega_j,b_j \in \Omega_k,i,j\in R^+$. Motivated by Lemma 2.2.3 with $\alpha_1=\psi_{m,d}^{\text{inf}}(\Omega_j)$ and $\alpha_2=\psi_{m,d}^{\text{inf}} (\Omega_k)$, for a given $n$ units, let ${X_{1:n}^{(i)}} \cap \Omega_j$ and ${X_{1:n}^{(i)}} \setminus{\Omega_j}$ be configurations of $N_j:=\left \lfloor \tilde{p}\cdot n \right \rfloor$ and $N_k:=n-N_j$ points, respectively, where
\begin{equation}\label{eqn:eqlab5el2adfg}
\tilde{p}=\frac{\psi_{m,d}^{\text{inf}} (\Omega_k)^{-\alpha_3}}{\psi_{m,d}^{\text{inf}} (\Omega_j)^{-\alpha_3}+\psi_{m,d}^{\text{inf}} (\Omega_k)^{-\alpha_3}},
\end{equation}
and $\left \lfloor x \right \rfloor$ is the floor function of $x$.
Let $X_{j,k}=X_{1:n}^{\Omega_j}\cup X_{1:n}^{\Omega_k}$, from the measure theory \cite{borodachov2019discrete,hunter2011measure} the following inequalities hold:
\begin{equation}
\label{eqn:eqlabel233a2a}
\begin{split}
\mathcal{E}^m_{\beta} (\Omega_j\cup \Omega_k) &\geq \mathcal{E}^m_{\beta}(X_{j,k}) \geq \min\left\{ \underset{x\in \Omega_j}{\inf} \underset{y\in X_{j,k}}{\sum}\frac{ \omega(x,y)}{\left \| x-y \right \|}, \underset{x\in \Omega_k}{\inf} \underset{y\in X_{j,k}}{\sum}\frac{ \omega(x,y)}{\left \| x-y \right \|}   \right\} \\
&\geq \min\left\{ \underset{x\in \Omega_j}{\inf} \underset{y\in X_{1:N}^{\Omega_j}}{\sum}\frac{ \omega(x,y)}{\left \| x-y \right \|}, \underset{x\in \Omega_k}{\inf} \underset{y\in X_{1:N}^{\Omega_k}}{\sum}\frac{ \omega(x,y)}{\left \| x-y \right \|}   \right\} \\
&=\min\left\{ \mathcal{E}^m_{\beta} (\Omega_j),\mathcal{E}^m_{\beta} (\Omega_k)  \right\}.
\end{split}
\end{equation}
Thus,
\begin{equation}
\label{eqn:eql25a}
\begin{split}
\psi_{m,d}^{\text{inf}} (\Omega_j \cup \Omega_k)&=\underset{n\to \infty}{\lim\inf}\frac{\mathcal{E}^m_{\beta} (\Omega_j\cup \Omega_k )}{n^{\frac{m}{d}+1}} \\
&\geq \underset{n\to \infty}{\lim\inf} \min\left\{ (\frac{N_j}{n})^{\frac{m}{d}+1}\cdot \frac{\mathcal{E}^m_{\beta} (\Omega_j) }{{N_j}^{\frac{m}{d}+1}},(\frac{N_k}{n})^{\frac{m}{d}+1}\cdot \frac{\mathcal{E}^m_{\beta} (\Omega_k) }{{N_k}^{\frac{m}{d}+1}}\right\} \\
&\geq \min\left\{ \tilde{p}^{\frac{m}{d}+1}\psi_{m,d}^{\text{inf}} (\Omega_j),(1-\tilde{p})^{\frac{m}{d}+1}\psi_{m,d}^{\text{inf}} (\Omega_k)  \right\}.
\end{split}
\end{equation}
When $\psi_{m,d}^{\text{inf}} (\Omega_j)=\psi_{m,d}^{\text{inf}} (\Omega_k)=\infty$, ~\eqref{eqn:eql25a} implies that $\psi_{m,d}^{\text{inf}} (\Omega_j \cup \Omega_k)=\infty$ and ~\eqref{eqn:eqlabel27dd} follows trivially. When $\psi_{m,d}^{\text{inf}} (\Omega_j)=\infty$ and $\psi_{m,d}^{\text{inf}} (\Omega_k)<\infty$, inequality ~\eqref{eqn:eql25a} becomes $\psi_{m,d}^{\text{inf}} (\Omega_j \cup \Omega_k)\geq (1-\tilde{p})^{\frac{m}{d}+1}\psi_{m,d}^{\text{inf}} (\Omega_k) $, we get $\psi_{m,d}^{\text{inf}} (\Omega_j \cup \Omega_k)\geq \psi_{m,d}^{\text{inf}} (\Omega_k)$ when $\tilde{p} \to 0$, which implies ~\eqref{eqn:eqlabel27dd}. When both $\psi_{m,d}^{\text{inf}} (\Omega_j)< \infty$ and $\psi_{m,d}^{\text{inf}} (\Omega_k) < \infty$, we deduce from ~\eqref{eqn:eql25a} and combine with ~\eqref{eqn:eqlab5el2adfg} we can get
\begin{equation*}\label{eqn:eqladbel27dd}
\psi_{m,d}^{\text{inf}} (\Omega_j \cup \Omega_k) \geq \left[ \psi_{m,d}^{\text{inf}} (\Omega_j)^{\alpha_3}+\psi_{m,d}^{\text{inf}} (\Omega_k)^{\alpha_3}
 \right]^\frac{1}{\alpha_3}.
\end{equation*}
Thus, Lemma 2.2.4 holds.

%(page 382 Lemma8.6.4) and page 394 Lemma8.7.3 
%Lemamma:11.1.5 page 482 
{\textbf{Lemma 2.2.5.}} $\exists \Omega_j,\Omega_k \subset \Omega$, and $\Omega_j,\Omega_k \not\subset  \emptyset $, $j\neq k$, $\mathcal{M}(d) > 1$ is continuous and derivative for  $d \in \mathbb{R}^+$, let $\alpha_3=\frac{1}{1-\mathcal{M}(d)}$ for $m >d$,
\begin{equation}\label{eqn:eqlabe33l27dd}
\psi_{m,d}^{\text{sup}} (\Omega_j \cup \Omega_k)^{\alpha_3} \ge \psi_{m,d}^{\text{sup}} (\Omega_j)^{\alpha_3}+\psi_{m,d}^{\text{sup}} (\Omega_k)^{\alpha_3}.
\end{equation}
Furthermore, if $\psi_{m,d}^{{\text{sup}}} (\Omega_j),\psi_{m,d}^{{\text{sup}}} (\Omega_k) \ge 0$ and at least one of these oracles is finite, then for any infinite subset $N'$ of $\mathbb{N}$ and any sequence $\{X_{1:n}\}_{n\in N'}$ of $n$-point configurations in $\Omega_j\cup \Omega_k$ such that
\begin{equation}\label{eqn:eqlabelx7}
\lim_{n\rightarrow \infty }\frac{\mathcal{E}^m_{\beta} (X_{1:n})}{\lambda (n)}=(\psi_{m,d}^{\text{sup}} (\Omega_j)^{\alpha_3}+\psi_{m,d}^{\text{sup}} (\Omega_k)^{\alpha_3})^\frac{1}{\alpha_3}
\end{equation}
holds, for a given $n$ units, let ${X_{1:n}^{(i)}} \cap \Omega_j$ and ${X_{1:n}^{(i)}} \setminus{\Omega_j}$ be configurations of $N_j:=\left \lfloor {p_j}\cdot n \right \rfloor$ and $N_k:=n-N_j$ points, respectively, we have
\begin{equation}\label{eqn:eqlab6}
p_j=\frac{\psi_{m,d}^{\text{sup}} (\Omega_j)^{\alpha_3}}{\psi_{m,d}^{\text{sup}} (\Omega_j)^{\alpha_3}+\psi_{m,d}^{\text{sup}} (\Omega_k)^{\alpha_3}}.
\end{equation}
$\textbf{Proof}$ When $\psi_{m,d}^{\text{sup}} (\Omega_j)=\infty$ and $\psi_{m,d}^{\text{sup}} (\Omega_k)<\infty$, $p_j$ would be $0$, while when $\psi_{m,d}^{\text{sup}} (\Omega_j)<\infty$ and $\psi_{m,d}^{\text{sup}} (\Omega_k)= \infty$, $p_j$ would be $1$. We assume $\psi_{m,d}^{\text{inf}} (\Omega)$ is bounded on the compact set $\Omega_j$ and $\Omega_k$. Let $\hat{N}$ be an infinite subsequence such that
\begin{equation*}\label{eqn:eqlabel24a}
\lim_{n\rightarrow \hat{N} }\frac{\mathcal{E}^m_{\beta} (\Omega_j \cup \Omega_k)}{n^{\mathcal{M}(d)}}:=\psi_{m,d}^{\text{inf}} (\Omega_j \cup \Omega_k).
\end{equation*}
Then, inspired by \cite{borodachov2019discrete,hunter2011measure}, for any $n\in N'$,
\iffalse
\begin{align*}
p(n)=
&\frac{1}{\pi \sqrt{2}} \sum_{k=1}^\infty \sqrt{k} A_k(n)
\frac{d}{dn} \Biggl\{{\frac {1} {\sqrt{n-\frac{1}{24}}} \\
&\times \sinh \Biggl[ {\frac{\pi}{k}
    \sqrt{\frac{2}{3}\Biggl(n-\frac{1}{24}\Biggr)}}\Biggr]
}\Biggr\}
\end{align*}
\fi
\begin{align*}
\label{eqn:eqlabel252a}
\begin{split}
&\mathcal{E}^m_{\beta} (X_{1:n})= \min\left\{ \underset{x\in \Omega_j}{\inf} \underset{y\in X_{1:n}}{\sum}\frac{ \omega(x,y)}{\left \| x-y \right \|}, \underset{x\in \Omega_k}{\inf} \underset{y\in X_{1:n}}{\sum}\frac{ \omega(x,y)}{\left \| x-y \right \|}   \right\}\\
&= \min \Biggl\{  \underset{x\in \Omega_j}{\inf} \left(  \underset{y\in {X_{1:n}} \cap \Omega_j}{\sum}\frac{ \omega(x,y)}{\left \| x-y \right \|}+\underset{y\in {X_{1:n}} \cap \Omega_k}{\sum}\frac{ \omega(x,y)}{\left \| x-y \right \|} \right),\\
&\ \ \ \ \ \ \ \ \ \ \ \ \ \ \ \underset{x\in \Omega_k}{\inf} \left(  \underset{y\in {X_{1:n}} \cap \Omega_j}{\sum}\frac{ \omega(x,y)}{\left \| x-y \right \|}+\underset{y\in {X_{1:n}} \cap \Omega_k}{\sum}\frac{ \omega(x,y)}{\left \| x-y \right \|} \right) \Biggr\}\\
&\le \min\left\{ \underset{x\in \Omega_j}{\inf} \underset{y\in X_{1:n}}{\sum}\frac{ \omega(x,y)}{\left \| x-y \right \|}, \underset{x\in \Omega_k}{\inf} \underset{y\in X_{1:n}}{\sum}\frac{ \omega(x,y)}{\left \| x-y \right \|}   \right\} +n \cdot \left\| \omega(x,y) \right\| \\
&\le \min\left\{ \mathcal{E}^m_{\beta} (\Omega_j),\mathcal{E}^m_{\beta} (\Omega_k)  \right\}+n \cdot \left\| \omega(x,y) \right\|.
\end{split}
\end{align*}
Thus,
\begin{equation}
\label{eqn:eql4525a}
\begin{split}
\psi_{m,d}^{\text{sup}} (\Omega_j \cup \Omega_k)&=\underset{n\to \infty}{\lim\sup}\frac{\mathcal{E}^m_{\beta} (\Omega_j\cup \Omega_k )}{n^{\frac{m}{d}+1}}\\
&\geq \underset{n\to \infty}{\lim\sup} \min\left\{ (\frac{N_j}{n})^{\frac{m}{d}+1}\cdot \frac{\mathcal{E}^m_{\beta} (\Omega_j) }{{N_j}^{\frac{m}{d}+1}},(\frac{N_k}{n})^{\frac{m}{d}+1}\cdot \frac{\mathcal{E}^m_{\beta} (\Omega_k) }{{N_k}^{\frac{m}{d}+1}}\right\}.
\end{split}
\end{equation}
When $\psi_{m,d}^{\text{sup}} (\Omega_j)< \infty$ and $\psi_{m,d}^{\text{sup}} (\Omega_k)= \infty$ from ~\eqref{eqn:eql4525a}, it follows that
\begin{equation}
\label{eqn:eql625a}
\begin{split}
\psi_{m,d}^{\text{sup}} (\Omega_j \cup \Omega_k)&\leq \underset{n\to \infty}{\lim\sup}  (\frac{N_j}{n})^{\frac{m}{d}+1}\cdot \frac{\mathcal{E}^m_{\beta} (\Omega_j) }{{N_j}^{\frac{m}{d}+1}}\le p_j^{-\alpha_3+1}\cdot \psi_{m,d}^{\text{sup}} (\Omega_j) \le  \psi_{m,d}^{\text{sup}} (\Omega_j)\\
&=\left[ \psi_{m,d}^{\text{sup}} (\Omega_j)^{\alpha_3}+\psi_{m,d}^{\text{sup}} (\Omega_k)^{\alpha_3} \right]^\frac{1}{\alpha_3}.
\end{split}
\end{equation}
Similarly, when $\psi_{m,d}^{\text{sup}} (\Omega_j)= \infty$ and $\psi_{m,d}^{\text{sup}} (\Omega_k)< \infty$, we obtain
\begin{equation}
\label{eqn:eql62555a}
\begin{split}
&\psi_{m,d}^{\text{sup}} (\Omega_j \cup \Omega_k)\le (1-p_j)^{-\alpha_3+1}\cdot \psi_{m,d}^{\text{sup}} (\Omega_j) \le  \psi_{m,d}^{\text{sup}} (\Omega_j)=\left[ \psi_{m,d}^{\text{sup}} (\Omega_j)^{\alpha_3}+\psi_{m,d}^{\text{sup}} (\Omega_k)^{\alpha_3} \right]^\frac{1}{\alpha_3}.
\end{split}
\end{equation}
Thus, both ~\eqref{eqn:eql625a} and ~\eqref{eqn:eql62555a} imply ~\eqref{eqn:eqlabel27dd}. 

When $\psi_{m,d}^{\text{sup}} (\Omega_j)< \infty$ and $\psi_{m,d}^{\text{sup}} (\Omega_k)< \infty$, ~\eqref{eqn:eql4525a} can be rewritten into
\begin{equation}
\label{eqn:eql623555a}
\begin{split}
\psi_{m,d}^{\text{sup}} (\Omega_j \cup \Omega_k)\le \min\left\{ p_j^{\frac{m}{d}+1}\psi_{m,d}^{\text{sup}} (\Omega_j),(1-p_j)^{\frac{m}{d}+1} \psi_{m,d}^{\text{sup}} (\Omega_k)\right\}.
\end{split}
\end{equation}
From Lemma 2.2.3, for the bounded $N_j$ and $N_k$, it follows that
\begin{equation*}\label{eqn:eqldabel27dd}
\psi_{m,d}^{\text{sup}} (\Omega_j \cup \Omega_k)^{\alpha_3} \ge \psi_{m,d}^{\text{sup}} (\Omega_j)^{\alpha_3}+\psi_{m,d}^{\text{sup}} (\Omega_k)^{\alpha_3}.
\end{equation*}
Thus, ~\eqref{eqn:eqlabel27dd} holds.
Combining ~\eqref{eqn:eqlabel27dd} with ~\eqref{eqn:eqlabelx7}, we get
\begin{equation}\label{eqn:eqlabe4lx7}
\lim_{n\rightarrow \infty }\frac{\mathcal{E}^m_{\beta} (\Omega_j \cup \Omega_k)}{\lambda (n)}=\psi_{m,d}^{\text{sup}} (\Omega_j \cup \Omega_k)^{\alpha_3}.
\end{equation}
Assume both $\psi_{m,d}^{\text{sup}} (\Omega_j)$ and $\psi_{m,d}^{\text{sup}} (\Omega_k)$ are finite, from ~\eqref{eqn:eqlabelx7} and ~\eqref{eqn:eql623555a} with Lemma 2.2.3,
\begin{equation}
\label{eqn:eql6235a}
\begin{split}
&\left[ \psi_{m,d}^{\text{sup}} (\Omega_j)^{\alpha_3}+\psi_{m,d}^{\text{sup}} (\Omega_k)^{\alpha_3} \right]^\frac{1}{\alpha_3}=\lim_{n\rightarrow \infty }\frac{\mathcal{E}^m_{\beta} (\Omega_j \cup \Omega_k)}{\lambda (n)}\\
&\le \min\left\{ p_j^{\frac{m}{d}+1}\psi_{m,d}^{\text{sup}} (\Omega_j),(1-p_j)^{\frac{m}{d}+1} \psi_{m,d}^{\text{sup}} (\Omega_k)\right\}\\
&\le \left[ \psi_{m,d}^{\text{sup}} (\Omega_j)^{\alpha_3}+\psi_{m,d}^{\text{sup}} (\Omega_k)^{\alpha_3} \right]^\frac{1}{\alpha_3}.
\end{split}
\end{equation}
From Lemma 2.2.3, we obtain
\begin{equation*}\label{eqn:eqlabd6}
p_j=\frac{\psi_{m,d}^{\text{sup}} (\Omega_j)^{\alpha_3}}{\psi_{m,d}^{\text{sup}} (\Omega_j)^{\alpha_3}+\psi_{m,d}^{\text{sup}} (\Omega_k)^{\alpha_3}},
\end{equation*}
as claimed in ~\eqref{eqn:eqlab6}. When $\psi_{m,d}^{\text{sup}} (\Omega_j)<\infty$ and $\psi_{m,d}^{\text{sup}} (\Omega_k)^{\alpha_3}=\infty$, from
~\eqref{eqn:eql625a}, we have
\begin{equation}
\label{eqn:eql62dd35a}
\begin{split}
\psi_{m,d}^{\text{sup}} (\Omega_j)&=\left[ \psi_{m,d}^{\text{sup}} (\Omega_j)^{\alpha_3}+\psi_{m,d}^{\text{sup}} (\Omega_k)^{\alpha_3} \right]^\frac{1}{\alpha_3}=\lim_{n\rightarrow \infty }\frac{\mathcal{E}^m_{\beta} (\Omega_j \cup \Omega_k)}{\lambda (n)}\\
&\le p_j^{\frac{m}{d}+1}\psi_{m,d}^{\text{sup}} (\Omega_j) \le  \psi_{m,d}^{\text{sup}} (\Omega_j),
\end{split}
\end{equation}
which can only be held if
\begin{equation}\label{eqn:eqldab6}
p_j=\frac{\psi_{m,d}^{\text{sup}} (\Omega_j)^{\alpha_3}}{\psi_{m,d}^{\text{sup}} (\Omega_j)^{\alpha_3}+\psi_{m,d}^{\text{sup}} (\Omega_k)^{\alpha_3}}=1.
\end{equation}
Similarly, when $\psi_{m,d}^{\text{sup}} (\Omega_j)=\infty$ and $\psi_{m,d}^{\text{sup}} (\Omega_k)^{\alpha_3}<\infty$, we obtain
\begin{equation}\label{eqn:eqldddab6}
p_j=\frac{\psi_{m,d}^{\text{sup}} (\Omega_j)^{\alpha_3}}{\psi_{m,d}^{\text{sup}} (\Omega_j)^{\alpha_3}+\psi_{m,d}^{\text{sup}} (\Omega_k)^{\alpha_3}}=0.
\end{equation}
Thus, ~\eqref{eqn:eqlab6} holds.

{\textbf{Lemma 2.2.6.}} Suppose that $m>d$, and $\Omega \subset \mathbb{R}^d$ is a compact set with $0<\mu(\Omega)<\infty$, $\mathcal{M}(d) > 1$ is continuous and derivative for  $d \in \mathbb{R}^+$. Furthermore, suppose that for any compact subset $\Omega_{i} \subset \Omega$, the limit $\psi_{m,d}(\Omega_{i}), i \in \mathbb{R}^+$ exists and is given by 
\begin{equation*}\label{eqn:eqlabelaa2a}
\psi_{m,d}(\Omega_{i})=\frac{{ \color{black}C_6}}{\mathcal{U}_d^m(\Omega_{i})^{\mathcal{M}(d)-1}}.
\end{equation*}
Then, $\psi_{m,d}(\Omega)$ exists and is given by
\begin{equation}\label{eqn:eqlabelaa2a2}
\psi_{m,d}(\Omega)=\frac{{ \color{black}C_6}}{\mathcal{U}_d^m(\Omega)^{\mathcal{M}(d)-1}}.
\end{equation}
Moreover, if a sequence of $n$-point configurations $X_{1:n}$ is asymptotically weighted Riesz polarization maximizing on the set $\Omega$ and $\mu(\Omega)>0$, then
\begin{equation}\label{eqn:eqlabel1a4}
\lim_{n\rightarrow \infty }\frac{1}{n}\sum_{i=1,i\neq j}^{n }\parallel x_i-x_j\parallel \rightarrow u_d^m(\mathbb{S}).
\end{equation}
{\textbf{Proof}} To prove ~\eqref{eqn:eqlabelaa2a2}, we firstly decompose the entire metric space $\Omega$ into extremely small disconnected parts with diameter less than $\epsilon>0$, according to the property of Borel metrics, then
\begin{equation}\label{eqn:eqlabel1a4sa}
\sum_{P\in \Omega_{i}}\mathcal{U}_d(P)\leq \mathcal{U}_d(\Omega).
\end{equation}
 Hereafter we follow an argument in \cite{borodachov2019discrete}, we define a sufficiently small space $\Omega_{i}$ as follows.
We consider the hyperplane $\Omega'$ consisting of all points, $(-l,l)$ is a cube embedded in $\Omega'$, we discretize the cube with tiny intervals for $j$-th ordinate, $-l= h_0^j<h_1^j\cdots <h_k^j= l,j\in i^{d'},d'\in(1,d),i=(i_1,i_2\cdots,i_N)$, $k$ is sufficiently large, $\exists \left \| h_k^j-h_{k-1}^j \right \|< \epsilon$ such that ~\eqref{eqn:eqlabel1a4sa} holds. $\Omega_{i}$ can be written as
\begin{equation*}\label{eqn:eqlabel1a4asa}
\Omega_{i}:=[ h_{i_1^1}^1,h_{i_1^1+1}^1)\times \cdots \times[h_{i_N^{d'}-1}^{d'},h_{i_N^{d'}}^{d'}).
\end{equation*}
For $\Omega_{i}\subset \Omega$, if $\omega (x_i,x_j)$ is bounded, let 
\begin{equation*}\label{eqn:eqlabel1a4as7a}
\overline{\omega}_{{\Omega}_{i}}=\sup_{x_i,x_j \in \Omega_{i}}\omega(x_i,x_j), \text{ and} \ \underline{\omega}_{{\Omega}_{i}}=\inf_{x_i,x_j \in \Omega_{i}}\omega(x_i,x_j),
\end{equation*}
we introduce the radial basis functions $\varphi(\cdot)$ to approximate the corresponding bounded $\omega(x_i,x_j)$:

\begin{align}
\label{eqn:eqlabel1a4asa2a}
\begin{split}
   &\overline{\omega}_{{\Omega}_{i}}(x_i,x_j)=\sum_{P\in \Omega_{i}}\overline{\omega}_P\varphi (\left \| x_i-x_j \right \|),\\ &\underline{\omega}_{{\Omega}_{i}}(x_i,x_j)=\sum_{P\in \Omega_{i}}\underline{\omega}_P\varphi (\left \| x_i-x_j \right \|).
\end{split}
\end{align}
From Lemma 2.2.5, and ~\eqref{eqn:eqlabe33l27dd}, there exists a ${ \color{black}C_6}$ satisfying
\begin{align*}
\label{eqn:eqlabel2aa}
\begin{split}
\psi_{m,d}^{{\text{sup}}} (\Omega)^{\alpha3} &\geq \sum_{i=1}^n\psi_{m,d}^{{\text{sup}}} (\Omega_i)^{\alpha3}\geq \sum_{i=1}^n\left [ \overline{\omega}_{{\Omega}_{i}}(x_i,x_j)\cdot \psi_{m,d}^{{\text{sup}}} (\Omega_i) \right ]^{\alpha3}={{ \color{black}C_6}}\sum_{x_i,x_j\in \Omega_{i}}\overline{\omega}_{{\Omega}_{i}}^{\alpha3}\cdot \mathcal{U}_d(\Omega_i)\\
&\geq {{ \color{black}C_6}}\int_{x_i,x_j \in \Omega_{i}}\overline{\omega}_{{\Omega}_{i}}(x_i,x_j)^{\alpha3}d\mathcal{U}_d(\Omega_i).
\end{split}
\end{align*}
%refer to Page 485 Chapter 9
From Lemma 2.2.4 and ~\eqref{eqn:eqlabel27dd}, similarly, we have
\begin{align*}
%\label{eqn:eqlabel2a3a}
\begin{split}
\psi_{m,d}^{{\text{inf}}} (\Omega)^{\alpha3} &\leq \sum_{i=1}^n\psi_{m,d}^{{\text{inf}}} (\Omega_i)^{\alpha3}\leq \sum_{i=1}^n\left [ \underline{\omega}_{{\Omega}_{i}}(x_i,x_j)\cdot \psi_{m,d}^{{\text{inf}}} (\Omega_i) \right ]^{\alpha3}={{ \color{black}C_6}}\sum_{x_i,x_j\in \Omega_{i}}\underline{\omega}_{{\Omega}_{i}}^{\alpha3}\cdot \mathcal{U}_d(\Omega_i)\\
&\leq {{ \color{black}C_6}}\int_{x_i,x_j \in \Omega_{i}}\underline{\omega}_{{\Omega}_{i}}(x_i,x_j)^{\alpha3}d\mathcal{U}_d(\Omega_i).
\end{split}
\end{align*}
Given a sufficiently small $P$, for ~\eqref{eqn:eqlabel1a4asa2a}, use the equation limit, we have
\begin{align*}
%\label{eqn:eqlabel2a3ax}
\begin{split}
& \overline{\omega}_{{\Omega}_{i}}(x_i,x_j)=\sum_{P\in \Omega_{i}}\overline{\omega}_P\varphi (\left \| x_i-x_j \right \|)={\omega}_{{\Omega}}(x_i,x_j),\\
&\underline{\omega}_{{\Omega}_{i}}(x_i,x_j)=\sum_{P\in \Omega_{i}}\underline{\omega}_P\varphi (\left \| x_i-x_j \right \|)={\omega}_{{\Omega}}(x_i,x_j).
\end{split}
\end{align*}
Since $\omega(x_i,x_j)$ is continuous on $\Omega$, both $\int_{x_i,x_j \in \Omega_{i}}\overline{\omega}_{{\Omega}_{i}}(x_i,x_j)^{\alpha3}d\mathcal{U}_d(\Omega_i)$ and $\int_{x_i,x_j \in \Omega_{i}}\underline{\omega}_{{\Omega}_{i}}(x_i,x_j)^{\alpha3}d\mathcal{U}_d(\Omega_i)$ converge to $\mathcal{U}_d^m(\Omega)$. Consequently, 
 the limit $\psi_{m,d}(\Omega_{i}), i \in \mathbb{R}^+$ exists and can be given by 
\begin{equation*}
\psi_{m,d}(\Omega_{i})=\frac{{ \color{black}C_6}}{\mathcal{U}_d^m(\Omega_{i})^{\mathcal{M}(d)-1}}.
\end{equation*}
By the Fatou's Lemma and Monotone Convergence Theorem, 
Thus,  ~\eqref{eqn:eqlabelaa2a2} holds on $\Omega$.

To prove ~\eqref{eqn:eqlabel1a4}, suppose that $X_{1:n}$ is an asymptotically weighted Riesz polarization maximizing sequence of $n$-point configuration on $\Omega$, the corresponding signed finite Borel measures $\cup_{i=1}^n{\mu_d^m(\Omega_i)}$ in $\mathbb{R}^d$ converges weak$^*$ to a signed finite Borel measure $\mu_d(\Omega)$, as $n \rightarrow \infty$. Consequently, ~\eqref{eqn:eqlabel1a4} is equivalent to the assertion that 
\begin{equation*}\label{eqn:eqlabel1a4a2aa}
\lim_{n\rightarrow \infty}\sum_{j=1}^{n}p_j=\cup_{i=1}^n \mu_d^m(\Omega_j)=\mu_d(\mathbb{S})
\end{equation*}
holds for any almost $\sigma\text{-algebra}$ subset on $\Omega$, let $\Omega_{\sigma}= \cup_{i=1}^n \Omega_i$ be a subset of $\sigma\text{-algebra}$ on $\Omega$,
\iffalse
According to the theorem 1.6.5 \cite{borodachov2019discrete}, we can infer 
\begin{equation}\label{eqn:eqlabel1aa4}
\lim_{N\rightarrow \infty}\frac{\text{Size}(X_{1:n} \cap A)}{n}=\mu_d(A)
\end{equation}

\fi
for any Borel subset $\Omega_{\sigma}\subset \Omega$. Since $\Omega_{\sigma}$ and $\Omega/\Omega_{\sigma}$ are the compact subsets of $\Omega$, suppose
$\psi_{m,d}(\Omega_{\sigma})=\frac{{ \color{black}C_7}}{\mu(\Omega_{\sigma})^{-\frac{1}{\alpha_3}}}$ and $\psi_{m,d}(\Omega/\Omega_{\sigma})=\frac{{ \color{black}C_7}}{\mu(\Omega/\Omega_{\sigma})^{-\frac{1}{\alpha_3}}}$, for the asymptotically weighted Riesz polarization maximal sequence $X_{1:n}$, 
\begin{align*}
%\label{eqn:eqlabel10huj3}
\begin{split}
 \lim_{n\rightarrow \infty }\frac{E^m (X_{1:n})}{\lambda (n)}& = { \color{black}C_7} \cdot (\mu(\Omega))^{\frac{1}{\alpha_3}} =  { \color{black}C_7} \cdot (\mu(\Omega_{\sigma})+\mu(\Omega/\Omega_{\sigma}))^{\frac{1}{\alpha_3}} \\
  & = \left [ \psi_{m,d}(\Omega_{\sigma})^{\alpha_3}+\psi_{m,d}(\Omega/\Omega_{\sigma})^{\alpha_3})\right ]^\frac{1}{\alpha_3}.
\end{split}
\end{align*}
Using ~\eqref{eqn:eqlabe33l27dd} in Lemma 2.2.5 and ~\eqref{eqn:eqlabelaa2a2} which holds for $\Omega_{\sigma}$ and $\Omega/\Omega_{\sigma}$, we have
\begin{align*}
\label{eqn:eqlabel10x32}
\begin{split}
 \lim_{n\rightarrow \infty}\sum_{j=1}^{n}p_j& =\frac{\psi_{m,d}(\Omega/\Omega_{\sigma})^{-\alpha_3}}{\psi_{m,d}(\Omega_{\sigma})^{-\alpha_3}+\psi_{m,d}(\Omega/\Omega_{\sigma})^{-\alpha_3}}=\frac{\mathcal{U}_d^m(\Omega_{\sigma})}{\mathcal{U}_d^m(\Omega_{\sigma})+\mathcal{U}_d^m(\Omega/\Omega_{\sigma})}=\mu_d^m(\mathbb{S}).
\end{split}
\end{align*}

Thus, ~\eqref{eqn:eqlabel1a4} holds.
\subsection{Bounds and Asymptotics }
To derive the asymptotics, we will first provide the lower and upper estimates of the Borel measure in a restricted compact space for the maximum weighted Riesz polarization quantity and then prove the asymptotics on $\mathbb{S}^2$.

{\textbf {Theorem 2.3.1.}} If $\Omega\subset \mathbb{R}^d$ is an infinite compact set, then
\begin{equation}\label{eqn:eqlabe34lx7}
\mathcal{E}^m_{\beta} (\Omega)\ge \frac{1}{n-1}\cdot \min_{x_i,x_j}\left \{ \sum_{i=1}^{n-1}\sum_{j=i+1}^{n}\frac{1}{\parallel x_i-x_j\parallel^{m+1}}  \right \}.
\end{equation}
If $\mathcal{U}_d^m(\mathbb{S})>0$, then there exists a constant ${ \color{black}C_8}>0$ depending only on $m$ such that
\begin{equation}\label{eqn:eqlabe334lx47}
\mathcal{E}^m_{\beta} (\Omega)\le \frac{{ \color{black}C_8}}{m-d+1}n^{\frac{m+1}{d}}.
\end{equation}
{\textbf{Proof}} Inspired by \cite{farkas2008transfinite}, define
\begin{equation}\label{eqn:e4334lx47}
D_n(\Omega)=\underset{X_1,\cdots,X_n\in\Omega}{\min}\frac{1}{n(n-1)}\sum_{i=1}^{n}\sum_{j=1,j\neq i}^{n}\frac{1}{\parallel x_i-x_j\parallel^{m+1}},
\end{equation}
we obtain \cite{farkas2008transfinite}
\begin{equation}\label{eqn:eqlabe34r447}
\sum_{i=1}^{n-1}\sum_{j=i+1}^{n}\frac{1}{\parallel x_i-x_j\parallel^{m+1}}\geq n \cdot D_n(\Omega).
\end{equation}
From the definition ~\eqref{eqn:eqlabe34r447} and ~\eqref{eqn:eqlabel1a4}, we have
\begin{equation}
\label{eqn:eqlabel233363a2a}
\begin{split}
\mathcal{E}_{\beta}^m (\Omega) &\geq \min_{x_i,x_j}\left \{ \sum_{i=1}^{n-1}\sum_{j=i+1}^{n}\frac{\omega(x_i,x_j)}{\parallel x_i-x_j\parallel^m}  \right \} \geq \min_{x_i,x_j}\left \{ \sum_{i=1}^{n-1}\sum_{j=i+1}^{n}\frac{1}{\parallel x_i-x_j\parallel^{m+1}}  \right \} \geq n \cdot D_n(\Omega) \\
&\geq \frac{1}{n-1}\cdot \min_{x_i,x_j}\left \{ \sum_{i=1}^{n-1}\sum_{j=i+1}^{n}\frac{1}{\parallel x_i-x_j\parallel^{m+1}}  \right \} .
\end{split}
\end{equation}
Thus, ~\eqref{eqn:eqlabe34lx7} holds.

Inspired by \cite{erdelyi2013riesz,kuijlaars1998asymptotics}, to prove ~\eqref{eqn:eqlabe334lx47}, let $X_{1:n}=\{x_1,...,x_n\}$ be a configuration of $n$ points on $\Omega$ that maximize the weighted Riesz polarization. Let $r_n= { \color{black}C_9} n^{-\frac{1}{d}}$, $\Omega_i:=\Omega\setminus{B(x,r_n)}$, where $B(x,r_n)$ is the open ball in $\mathbb{R}^d$ with center $x$ and radius $r_n$, we have 
\begin{equation}
\label{eqn:eqlabel2333634442a}
\begin{split}
&\mu(B(x,r_n)\cap \Omega)\le { \color{black}C_9}r_n^d,\\
&\mu(\Omega)\ge 1-\sum_{j=1}^{n}\mu(B(x_j,r_n))\ge 1-{ \color{black}C_9}nr_n^d.
\end{split}
\end{equation}
The inequality of the quantity $\mathcal{E}_{\beta}^m (\Omega)$ can be expressed by
\begin{equation}
\label{eqn:e42a}
\begin{split}
\mathcal{E}_{\beta}^m (\Omega)&\le \frac{\left\| \omega(x_i,x_j) \right\|}{\mu(\Omega)}\int_{\Omega_i}^{}\sum_{j=1}^{n}\left\| x-x_j \right\|^{-(m+1)}d\mu(x)\\
&\le \frac{1}{1-{ \color{black}C_9}nr_n^d}\sum_{j=1}^{n}\int_{\Omega_i}^{}\left\| x-x_j \right\|^{-(m+1)}d\mu(x).
\end{split}
\end{equation}
From \cite{erdelyi2013riesz}, we obtain the integral inequality 
\begin{equation}
\label{eqn:e45442a}
\begin{split}
\int_{\Omega_i}^{}\left\| x-x_j \right\|^{-(m+1)}d\mu(x)&=\int_{0}^{\infty}\mu\left\{ x\in\Omega_i:\left\| x-x_j \right\|^{-(m+1)}>t \right\} dt\\
&\le 1+\int_{1}^{r_n^{-(m+1)}}\mu(B(x_j,t^{-\frac{1}{m+1}})\cap \Omega) dt\\
&\le 1+{ \color{black}C_9}\int_{1}^{r_n^{-(m+1)}}t^{-\frac{d}{m+1}} dt,
\end{split}
\end{equation}
when $n$ is sufficiently large such that $r_n^{-(m+1)}>1$. For $m>d$, thus ~\eqref{eqn:e42a} follows that
\begin{equation}
\label{eqn:e42dda}
\begin{split}
\mathcal{E}_{\beta}^m (\Omega)
&\le \frac{n}{1-{ \color{black}C_9}nr_n^d}(1+{ \color{black}C_9}\int_{1}^{r_n^{-(m+1)}}t^{-\frac{d}{m+1}} dt)\le \frac{n}{1-{ \color{black}C_9}nr_n^d}(1+{ \color{black}C_9}\int_{1}^{r_n^{-(m+1)}}t^{-\frac{d}{m+1}} dt)\\
&= \frac{n}{1-{ \color{black}C_9}nr_n^d} \cdot \frac{(m+1)(r_n^{d-m-1}-1)}{m-d+1}\le \frac{{ \color{black}C_8}}{m-d+1}n^{\frac{m+1}{d}}.
\end{split}
\end{equation}
Thus, ~\eqref{eqn:eqlabe334lx47} holds.

Regarding the asymptotic behavior of the weighted Riesz polarization as $m$ approximate the $d$, we have

{\textbf {Theorem 2.3.2.}} For $m>d$, if $\Omega=\mathbb{S}^d$,
\begin{equation}\label{eqn:eqlabe34lx474d}
\lim_{m \to d^+}\underset{n \to \infty}{\lim\inf}\frac{\mathcal{E}^m_{\beta} (\Omega)}{n^{\frac{m+1}{d}}}=\infty.
\end{equation}
{\textbf{Proof}} From \cite{erdelyi2013riesz}, let $\mathcal{E}(\Omega):=\sum_{i=1}^{n}\sum_{j=i+1}^{n}\frac{1}{\left\| x_i-x_j \right\|^m}$, we have
\begin{equation}\label{eqn:eq677lx474d}
\underset{n \to \infty}{\lim\inf}\frac{\mathcal{E}(\Omega)}{n\log n}\ge \frac{c_d}{u_d(\mathbb{S}^d)}=\frac{\Gamma(\frac{d+1}{2})}{\sqrt{\pi}d\cdot\Gamma(\frac{d}{2})}:=\tau_d,
\end{equation}
when $m=d$. Let $\Omega=\mathbb{S}^d$, we have the estimate for $x\in \mathbb{S}^d$ \cite{kuijlaars1998asymptotics}
\begin{equation}\label{eqn:eq65674d}
\mu(B(x,r)\cap \mathbb{S}^d)\le \tau_dr^d,
\end{equation}
and for $m>d$, from the estimate
\begin{equation}
\label{eqn:e454464772a}
\begin{split}
\int_{\Omega\setminus{B(x_j,r_n)}}^{}\left\| x-x_j \right\|^{-(m+1)}d\mu(x)&=d \tau_d 2^{-\frac{m+1}{2}}\int_{-1}^{1-\frac{r_n^2}{2}}(1-t)^{-\frac{m+1}{2}+\frac{d}{2}-1}(1+t)^{\frac{d}{2}-1} dt\\
&\le d\tau_d 2^{-\frac{m+1}{2}+\frac{d}{2}-1}\int_{-1}^{1-\frac{r_n^2}{2}}(1-t)^{-\frac{m+1}{2}+\frac{d}{2}-1}dt\\
&=\frac{d\tau_d}{m+1-d}\left[ r^{-m-1+d}-2^{-p-1+d} \right],r<2 ,
\end{split}
\end{equation}
Substitute ~\eqref{eqn:e454464772a} into ~\eqref{eqn:e42a}, we have
\begin{equation}
\label{eqn:e42456dda}
\begin{split}
\mathcal{E}_{\beta}^m (\Omega)
&\le \frac{n}{1-{ \color{black}C_9}nr_n^d}\cdot \frac{d\tau_d}{m+1-d}\cdot r^{-m-1+d}.
\end{split}
\end{equation}
The optimal value for $r_n$ is
\begin{equation}
\label{eqn:e4245556dda}
\begin{split}
r_n=(\frac{m+1-d}{nm\tau_d+n\tau_d})^\frac{1}{d}.
\end{split}
\end{equation}
Substitute ~\eqref{eqn:e4245556dda} and ~\eqref{eqn:e42456dda} into ~\eqref{eqn:eqlabe34lx474d}, the inequality holds.

% covering and separation of chebychev  
%14.4
\subsection{ \color{black}Covering Radius }

%2011_Quasi-uniformity of minimal weighted energy points on compact metric space

  In this section, we state and prove the bound of the covering radius. And extend to deal with the weak* limit distribution of best-covering $n$-point configurations on rectifiable sets $\Omega$. Suppose that $\Omega$ is a compact infinite metric space with Euclidean metric $r(x,y)= \left\| x-y \right\|$, $\Omega \times \Omega \rightarrow [0,\infty )$, we define the covering radius of an $n$-point configuration $X_{1:n}$ in a metric space $(\Omega,r)$ as $\rho(X_{1:n},\Omega):= \max_{x \in \Omega}\min_{i=1,...,n}r(x,x_i).$
\iffalse
\begin{equation*}\label{eqn:eqlabel45b}
\rho(X_{1:N},\Omega):= \max_{x \in \Omega}\min_{i=1,...,N}r(x,x_i).
\end{equation*}
\fi
From the geometrical perspective, the covering radius of $X_{1:n}$ can be considered as the minimal radius of $n$ adjacent closed balls centered at $X_{1:n}$ whose union contains the entire $\Omega$. Among finite element analysis and approximation theory, this quantity is known as the best approximation of the set $\Omega$ by the configuration $X_{1:n}$ \cite{borodachov2019discrete}. The optimal values of this quantity are also of interest and we define the minimal $n$-point covering radius of a set $\Omega$ as
\begin{equation*}\label{46b}
%\rho_N(\Omega):=\rho_N^{\Omega}(\Omega)=\min\{\rho(X_{1:N},\Omega): X_{1:N} \subset \Omega\}.
\rho_n(\Omega):=\min\{\rho(X_{1:n},\Omega): X_{1:n} \subset \Omega\}.
\end{equation*}
$\rho_n(\Omega)$ is also called an $n$-point best-covering configuration for $\Omega$ \cite{hardin2012quasi}.
\iffalse
 Covering radius arises in the optimal configurations of a given compact finite space. We generally impose some restrictions on this structure and define the mesh ratio as
\begin{equation}\label{47b}
%\rho_N(\Omega):=\rho_N^{\Omega}(\Omega)=\min\{\rho(X_{1:N},\Omega): X_{1:N} \subset \Omega\}.
\tau(X_{1:N},\Omega)=\rho(X_{1:N},\Omega)/r_{\text{min}}.
\end{equation}
If $\tau(X_{1:N},\Omega)$ is bounded, the $N$-point configuration is said to be quasi-uniform on $\Omega$.\\
\fi
%{ \color{black}We start by following basic estimate for our energy criterion.}\\
\iffalse
\textbf{Proposition 6} Let $m>d>0$, there exists a constant $C_x>0$ such that for any compact  set $\Omega_i \subset R^d$ with the associated positive measure $\mu(\Omega_i^{\infty})$, and any $N$-point energy minimizing configuration $X_{1:N}$ on $\Omega$:
\begin{equation}\label{471b}
%\rho_N(\Omega):=\rho_N^{\Omega}(\Omega)=\min\{\rho(X_{1:N},\Omega): X_{1:N} \subset \Omega\}.
\rho(X_{1:N},\Omega) \geq C_x(\frac{N}{\mu(\Omega)})^{-m}.
\end{equation}\\
Proof: For any compact set, there is a positive Borel measure, then
\begin{equation}\label{472b}
%\rho_N(\Omega):=\rho_N^{\Omega}(\Omega)=\min\{\rho(X_{1:N},\Omega): X_{1:N} \subset \Omega\}.
r_N \geq (\frac{\mu(\Omega)}{N})^d \geq C_x \frac{\mu(\Omega_A^{\infty})}{\mu(\Omega)}N^{\phi(d)}.
\end{equation}
\fi

%Theory 2.6 in covering and separation peroperites of chebyshev 
{\textbf{Theorem 2.4.1.}} Suppose the compact set $\Omega \subset \mathbb{S}^d$ with $\mathcal{U}_d^m(\Omega)>0$, there exists a positive constant $ \color{black}C_{10}$ such that for any $n$
-point configuration $X_{1:n}^*$ that is optimal for $\mathcal{E}_{\beta} (\Omega)$, we have $\rho(X_{1:n}^*,\Omega)\le { \color{black}C_{10}} \cdot n^{-\frac{m-2d}{d\cdot (m-d)}}$, where ${ \color{black}C_{10}} \propto  \left( \frac{m}{m-d} \right)^{\frac{1}{m-d}}$.

{\textbf{Proof}} Since $\Omega\subset \mathbb{S}^d$ is a compact set, there exists a finite family of set $\left\{ \Omega_i \right\},i=1,...,n'$, with the following properties: $(1)$ $\Omega=\left\{ \cup \Omega_i \right\},i=1,...,n', $ and the interiors of the sets $\Omega_i$ are disjoint where the measure $\mu(\Omega_i\cap \Omega_j)_{i\neq j}=0$. $(2)$ There exist positive constants $ \color{black}C_{11}$ and $ \color{black}C_{12}$, that does not depend on $n$, and the point $x_i\in \Omega_i$, such that $B(x_i,{ \color{black}C_{11}}n^{-\frac{1}{d}-\frac{1}{m}})\cap \Omega\subset \Omega_i\subset B(x_i,{ \color{black}C_{12}}n^{-\frac{1}{d}-\frac{1}{m}})$. Since $\Omega_i \subset B(x_i,n^{-\frac{1}{d}-\frac{1}{m}})$, there exists a $\alpha'$ such that the number of points from $X_{1:n}^*$ is $\# (\Omega_i\cap X_{1:n}^*)\le \alpha'n$, where $ 0< \alpha'< 1$.

Hereafter we follow an argument in \cite{reznikov2018covering}. Let $y\in \Omega$ be such that $\underset{x_k\in X_{1:n}^*}{\min}\left| y-x_k \right|=\rho(X_{1:n},\Omega)$. Assume $\rho(X_{1:n},\Omega) \geq { \color{black}C_{13}}n^{-\frac{1}{d}-\frac{1}{m}}$, for every $x_i\in\left\{ \left\{ X_{1:n}^* \right\}\cap \Omega_i \right\}$, we have
\begin{equation}
\label{eqn4d4d}
\begin{split}
\left| y-x \right|&\le \left| y-x_i \right|+\left| x_i-x \right|\le \left| y-x_i \right|+2{ \color{black}C_{12}}n^{-\frac{1}{d}-\frac{1}{m}}\le \left| y-x_i \right|+\frac{2{ \color{black}C_{12}}}{{ \color{black}C_{13}}}\rho(X_{1:n},\Omega)\\
&\le \frac{2{ \color{black}C_{12}}+{ \color{black}C_{13}}}{{ \color{black}C_{13}}}\left| y-x_i \right|,
\end{split}
\end{equation}
which implies
\begin{equation}\label{eqn:eq67d4d}
\left| y-x_j \right|^{-m}\le \frac{2{ \color{black}C_{12}}+{ \color{black}C_{13}}}{{ \color{black}C_{13}}}\cdot \underset{x\in\Omega_i}{\min}\left| y-x \right|^{-m}.
\end{equation}
The corresponding lower bound
\begin{equation}
\label{eqn4dd4d}
\begin{split}
\left| y-x \right|&\ge \left| y-x_i \right|-\left| x_i-x \right|\ge \left| y-x_i \right|-2{ \color{black}C_{12}}n^{-\frac{1}{d}-\frac{1}{m}}\le \left| y-x_i \right|-\frac{{ \color{black}C_{12}}}{{ \color{black}C_{13}}}\rho(X_{1:n},\Omega)\\
&\ge \frac{{ \color{black}C_{13}}-{ \color{black}C_{12}}}{{ \color{black}C_{13}}}\rho(X_{1:n},\Omega).
\end{split}
\end{equation}
Consequently,
$\Omega\cap B(y,\frac{{ \color{black}C_{13}}-{ \color{black}C_{12}}}{{ \color{black}C_{13}}}\rho(X_{1:n},\Omega))\subset \Omega\setminus{\bigcup_{x_i\in X_{1:n}^*}\Omega_i}$.
For each $x_i\in\Omega_i$, from ~\eqref{eqn:eq67d4d}, we get
\begin{equation}\label{eqn:eq447d4d}
\frac{1}{\left| y-x_i \right|^m}\le \left( \frac{2{ \color{black}C_{12}}+{ \color{black}C_{13}}}{{ \color{black}C_{13}}} \right)^m\frac{1}{\mu(\Omega)}\int_{\Omega_i}\frac{d\mu(x)}{\left| y-x \right|^m}.
\end{equation}
Since $B(x_i,{ \color{black}C_{12}}n^{-\frac{1}{d}-\frac{1}{m}})\cap \Omega\subset \Omega_i$, we get $\mu(\Omega)\ge { \color{black}C_{12}}\cdot n^{-1}$, which implies
\begin{equation}
\label{eqnddda}
\begin{split}
\mathcal{E}^m_{\beta} (\Omega)&={ \color{black}C_{12}}\cdot n^{\frac{m}{d}+1}\le n \cdot \sum_{x_i\in X_{1:n}}\frac{1}{\left| y-x_i \right|^m}\le n \cdot \left( \frac{2{ \color{black}C_{12}}+{ \color{black}C_{13}}}{{ \color{black}C_{13}}} \right)^m \sum_{x_i \in X_{1:n}}\frac{1}{\mu(\Omega_i )}\int_{\Omega_i}\frac{d\mu(x)}{\left| y-x \right|^m}\\
&\le n^2 \left( \frac{2{ \color{black}C_{12}}+{ \color{black}C_{13}}}{{ \color{black}C_{13}}} \right)^m \underset{B(y,\frac{{ \color{black}C_{13}}-{ \color{black}C_{12}}}{{ \color{black}C_{13}}}\rho(X_{1:n},\Omega))}{\int }\frac{d\mu(x)}{\left| y-x \right|^m}\\
&\le \frac{m}{m-d} \cdot \left( \frac{2{ \color{black}C_{12}}+{ \color{black}C_{13}}}{{ \color{black}C_{13}}} \right)^m\cdot n^2\cdot \left( \frac{{ \color{black}C_{13}}-{ \color{black}C_{12}}}{{ \color{black}C_{13}}}\rho(X_{1:n},\Omega) \right)^{d-m},
\end{split}
\end{equation}
which implies
\begin{equation}\label{eqdd7d4d}
\left[ \rho(X_{1:n},\Omega) \right]^{m-d}\le \frac{m}{m-d}\cdot \left( \frac{2{ \color{black}C_{12}}+{ \color{black}C_{13}}}{{ \color{black}C_{13}}} \right)^m \cdot n^{-\frac{m-2d}{d}},
\end{equation}
we get
\begin{equation}\label{eqddd564d}
\rho(X_{1:n},\Omega) \le \left( \frac{m}{m-d} \right)^{\frac{1}{m-d}}\cdot \left( \frac{2{ \color{black}C_{12}}+{ \color{black}C_{13}}}{{ \color{black}C_{13}}} \right)^{\frac{m}{m-d}} \cdot n^{-\frac{m-2d}{d\cdot (m-d)}}.
\end{equation}

A good estimate on the constant ${ \color{black}C_{10}}$ for large values of $m$ yields the following theorem regarding the asymptotic behavior of $\mathcal{E}_{\beta} (\Omega)$ as $m \to \infty$.

{\textbf{Theorem 2.4.2.}} Suppose the compact set $\Omega \subset \mathbb{S}^d$ or $\Omega=[0,1]^d$. The quantities as defined in {\textbf {Theorem 2.2.1}}, the following limits exist as positive real numbers and satisfy
\begin{equation}
\label{eqnddddda}
\begin{split}
&\lim_{m\rightarrow \infty }\frac{{ \color{black}C_{2}}}{[\mathcal{U}_d^m(\Omega)]^{\frac{1}{d}}}=\lim_{m\rightarrow \infty }\left( \lim_{n\rightarrow \infty }\frac{\mathcal{E}_{\beta} (\Omega)}{n^{\frac{1}{m}+\frac{1}{d}}} \right)=\frac{1}{\lim_{n\rightarrow \infty}n^\frac{1}{d}\rho_n(\Omega)}.
\end{split}
\end{equation}
{\textbf{Proof}} If $A \subset \mathbb{S}^d$, there exist positive constant ${ \color{black}C_{11}}$ and ${ \color{black}C_{12}}$ such that ${ \color{black}C_{11}}n^{-\frac{1}{d}-\frac{1}{m}} \le \rho_n(\Omega)\le { \color{black}C_{12}}n^{-\frac{1}{d}-\frac{1}{m}} $. Observe that
\begin{equation}
\label{eqn67dda}
\begin{split}
\mathcal{E}_{\beta} (\Omega)&\ge \underset{y\in \Omega}{\inf}\sum_{x_i\in X_{1:n}^*}^{}\frac{1}{\left| y-x_i \right|^{\frac{1}{m}+1}}=\frac{1}{\underset{y\in \Omega}{\max}\underset{x_i \in X_{1:n}^*}{\min}{\left| y-x_i \right|^{\frac{1}{m}+1}}}=\rho_n(\Omega)^{-1-\frac{1}{m}}\ge { \color{black}C_{13}}n^{\frac{1}{d}+\frac{1}{m }}.
\end{split}
\end{equation}
Consequently,
\begin{equation}
\label{eqn6756a}
\begin{split}
\lim_{n\rightarrow \infty }\frac{\mathcal{E}_{\beta} (\Omega)}{n^{\frac{1}{m}+\frac{1}{d}}}\ge \frac{1}{\underset{n\to \infty}{\lim\inf}(n^{\frac{1}{d}+\frac{1}{m}}\rho_n(\Omega))},
\end{split}
\end{equation}
which implies
\begin{equation}
\label{eqn56a}
\begin{split}
\underset{m\to \infty}{\lim\inf}\left( \lim_{n\rightarrow \infty }\frac{\mathcal{E}_{\beta} (\Omega)}{n^{\frac{1}{m}+\frac{1}{d}}} \right)\ge \frac{1}{\underset{n\to \infty}{\lim\inf}(n^\frac{1}{d}\rho_n(\Omega))}. 
\end{split}
\end{equation}
Using the same argument as \cite{reznikov2018covering}, we now take an arbitrary point $y\in \Omega$ such that 
\begin{equation}
\label{eqn4556a}
\begin{split}
\underset{j=1,\cdots,n}{\min}\left| y-x_i \right|=\rho_n(X_{1:n}^*), 
\end{split}
\end{equation}
and set $B_i:=B(y,i\cdot \rho_n(X_{1:n}^*))\setminus B(y,(i-1)\cdot \rho_n(X_{1:n}^*))$, where $i\ge 2$. Since $X_{1:n}^*\cap B(y, \rho_n(X_{1:n}^*))=\emptyset$, we have $X_{1:n}^*\subset \bigcup_{i=2}^{\infty}B_i$. For any $i \ge 2$ we have
\begin{align}
\label{eqn:eq34bel10}
\begin{split}
 \mathcal{E}_{\beta} (\Omega)&=\max_{\Omega} \min_{x_i,x_j}\left \{ \sum_{i=1}^{n-1}\sum_{j=i+1}^{n}\frac{\omega(x_i,x_j)}{\parallel x_i-x_j\parallel^m} \right \}^{\frac{1}{m}}
\le \max_{\Omega}\left \{ \sum_{i=1}^{n-1}\sum_{j=i+1}^{n}\frac{\omega(x_i,x_j)}{\parallel x_i-x_j\parallel^m}  \right \}^{\frac{1}{m}}\\
 &\le \left|  \omega(x_i,x_j) \right|^{\frac{1}{m}}\max_{\Omega}\left \{ \sum_{i=1}^{n-1}\sum_{j=i+1}^{n}\frac{1}{\parallel x_i-x_j\parallel^m}  \right \}^{\frac{1}{m}}\\
 &\le \left|  \omega(x_i,x_j) \right|^{\frac{1}{m}}\max \left( \sum_{i=2}^{\infty} \sum_{x\in X_{1:n}^*\cap B_i}^{}\frac{1}{\left| y-x \right|^m} \right)^\frac{1}{m}.
 % & \omega (x_i,x_j)=\left [ \gamma (x_i) \gamma (x_j) +\beta \parallel x_i-x_j\parallel \right ]^{-\frac{m}{2d}}\omega' (x_i,x_j). \\
 %&  \omega' (x_i,x_j)=\Phi (\frac{\parallel x_i-x_j\parallel}{r_N}), \\
 %& r_N=C_NN^\frac{-1}{d}, \exists \beta, C_N\in R^+. \\
\end{split}
\end{align}
By the property of $B_n$, for any $x\in B_i$, we have $\left| y-x \right|\ge (i-1)\cdot\rho_n(X_{1:n}^*)$, which implies
\begin{align}
\label{eqn:eq35670}
\begin{split}
 \mathcal{E}_{\beta} (\Omega)\le \left|  \omega(x_i,x_j) \right|^{\frac{1}{m}}\cdot \left[ \sum_{i=2}^{\infty}i^d\cdot(i-1)^{-1} \right]^\frac{1}{m}\cdot\rho_n(X_{1:n}^*)^{-1}.
\end{split}
\end{align}
Dividing by $n^{\frac{1}{m}+\frac{1}{d}}$, since $\rho_n(X_{1:n}^*)\ge \rho_n(\Omega)$, we get
\begin{align}
\label{eqn:eq354570}
\begin{split}
 \frac{\mathcal{E}_{\beta} (\Omega)}{n^{\frac{1}{m}+\frac{1}{d}}}\le \frac{\left|  \omega(x_i,x_j) \right|^{\frac{1}{m}}\cdot \left[ \sum_{i=2}^{\infty}i^d\cdot(i-1)^{-1} \right]^\frac{1}{m}}{n^{\frac{1}{m}+\frac{1}{d}}\rho_n(\Omega)},
\end{split}
\end{align}
which implies
\begin{align}
\label{eqn:eq56090}
\begin{split}
 \lim_{n \to \infty}\frac{\mathcal{E}_{\beta} (\Omega)}{n^{\frac{1}{m}+\frac{1}{d}}} \le \frac{\left|  \omega(x_i,x_j) \right|^{\frac{1}{m}}\cdot \left[ \sum_{i=2}^{\infty}i^d\cdot(i-1)^{-1} \right]^\frac{1}{m}} {\underset{n \to \infty}{\lim\sup}\left( n^{\frac{1}{m}+\frac{1}{d}}\rho_n(\Omega) \right)}.
\end{split}
\end{align}
As $m \to \infty$, we have
\begin{align}
\label{eqn:eq56y90}
\begin{split}
 \underset{n \to \infty}{\lim\sup}\left( \lim_{n \to \infty}\frac{\mathcal{E}_{\beta} (\Omega)}{n^{\frac{1}{m}+\frac{1}{d}}} \right) \le \frac{1} {\underset{n \to \infty}{\lim\sup}\left( n^{\frac{1}{m}+\frac{1}{d}}\rho_n(\Omega) \right)}.
\end{split}
\end{align}
~\eqref{eqn56a} and ~\eqref{eqn:eq56y90} imply that $\underset{n \to \infty}{\lim\sup}\left( n^{\frac{1}{m}+\frac{1}{d}}\rho_n(\Omega) \right)$ and $\underset{n \to \infty}{\lim\sup}\left( \lim_{n \to \infty}\frac{\mathcal{E}_{\beta} (\Omega)}{n^{\frac{1}{m}+\frac{1}{d}}} \right)$ exist and satisfy
\begin{equation}
\label{eqnd45dda}
\begin{split}
\lim_{m\rightarrow \infty }\left( \lim_{n\rightarrow \infty }\frac{\mathcal{E}_{\beta} (\Omega)}{n^{\frac{1}{m}+\frac{1}{d}}} \right)
=\frac{1}{\lim_{n\rightarrow \infty}n^\frac{1}{d}\rho_n(\Omega)}.
\end{split}
\end{equation}
Thus, ~\eqref{eqnddddda} holds.
%{\textbf{Corollary 7.1}}
%\section{3 \ \ Polarization Inequalities via Stein Methods }

\section{ Weighted Chebyshev Particles MCMC}
%\section{Weighted Riesz $\beta$-energy Particles MCMC}
%2016-comparing two recent particle filter_PMH_SMC2
%2015-Sequential Exploration of Complex Surfaces Using Minimum Energy Designs.
%2017-Deterministic Sampling of Expensive Posteriors using minimum energy designs

In this section, we will develop a new sampler, where the propagation of particles is derived from weighted Riesz polarization maximizing, since this quantity inherits some properties of the Chebyshev constant, these samplers are called Chebychev Particles, the sample inherits the special features presented in Section 2 when traversing in a discretized deterministic submanifolds of parameter space via pairwise interactions. We further extend it to sequential sampling in the particle Metropolis-Hastings framework for the inference of hidden Markov models, where the acceptance ratio is approximated by a pseudo-marginal Metropolis-Hastings algorithm. 
\iffalse
\begin{algorithm}[t]
\SetAlgoLined
\KwResult{Weighted particles $\{X_{1:N}^i,W_{1:N}^i\}$ for targets}
 Draw $x_1^{i}\leftarrow q(x_1|y_1)=\text{arg}\min_{x}\gamma (x)$. \\
 Compute $w_1(x_1^i)=\frac{g(y_1|X_1^i)\cdot \gamma(x_1)}{q(x_1^i|y_1)},W_1^i \propto w_1(x_1^i)$.\\
 Normalize by $W_1^i \leftarrow \frac{W_1^i}{\sum_{i=1}^{N_p}W_1^i}$ to obtain updated weighted particles $\{X_1^i,W_1^i\}$ for $t=1$, then $t \leftarrow t+1$. \\
 \While{$t \leq T$}{
  Resample ${a_{t_1}^{(i)}}$ with $\mathbb{P}({a_{t-1}^{(i)}}=j)\leftarrow w_{t-1}^{(j)}.$ \\
  Draw $X_t^i\leftarrow q(x_t|y_t,X_{t-1}^{a_{t_1}^{(i)}})=\text{arg}\min_x \Omega_{\beta}(m,N)$, \; \\
  Update set $X_{1:t}^i\leftarrow (X_{1:t-1}^{a_{t_1}^{(i)}},X_t^i)$,\\
  Compute $w_t(X_{t-1:t}^i)=\frac{g(y_t|X_t^i)f(X_t^i|X_{t-1}^{a_{t_1}^{(i)}})}{q(x_{t}|y_t,X_{1:t-1}^{a_{t_1}^{(i)}})},W_t^i \propto w_t(X_{t-1:t}^{a_{t_1}^{(i)}})$.\\
  Normalize by $W_t^i \leftarrow \frac{W_t^i}{\sum_{i=1}^{N_p}W_t^i}$, and $t \leftarrow t+1$\\
  \iffalse
  \eIf{t==1}{
   instructions1\;
   instructions2\;
   }{
   instructions3\;
  }
  \fi
 }
\caption{Weighted Riesz Particles $\beta$-energy MCMC.}\label{algo:event2}
\end
{algorithm}
\fi
\iffalse
All these special features will be inherited when we embed weighted Riesz particles into MCMC, the new sampler conducts high performance, which is verified by experiment for hidden Markov models in section 4.

\fi
\subsection{Sequential Chebyshev Particles Sampling}

Finding the optimal designs of configurations is nondeterministic, especially for high dimensions, where point-by-point traversal results in exponential growth in computational load. A number of optimization algorithms were proposed for the optimal design of different configurations.  Park~\cite{park1994optimal} proposed a 2-stage exchange and Newton-type for optimal designs which minimize the integrated mean squared error and maximize entropy, respectively. Ye~\cite{ye1998orthogonal} further extended it by the column-pairwise algorithm. Morris and Mitchell~\cite{morris1995exploratory} adapted simulated annealing \cite{kirkpatrick1983optimization} to explore the unit in a reachable domain. %\cite{jin2003efficient} proposed an enhanced stochastic evolutionary (ESE) algorithm where warming schedule and cooling schedule jointly modulate the relevant parameters through two loops so that the algorithm can be adapted to different optimization design criteria. The inner loop selects new configurations from the neighborhood of the current configuration and decides whether to accept it based on an acceptance criterion. The outer loop modulates the entire optimization process by adjusting the threshold.
Inspired by \cite{morris1995exploratory} and \cite{joseph2015sequential}, we propose a constrained one-point-per-time greedy algorithm for developing the sequential designs of weighted Riesz particles as follows.

(\uppercase\expandafter{\romannumeral1}) The choice of the initial point is crucial since it is closely related to sampling the subsequent points. %If $x_0$ locates inside the compact space $\Omega$, the total configuration error generated by the convergence of the greedy algorithm is very small compared to the ground truth. If not, the configuration is from the points corresponding to the local optimum of the energy criterion, which cause serious numerical instability. Consequently, 
For the sake of numerical stability, we take the particle with the largest average value as the initial point. We have the expectation, $\mathbb{E}
(x)=\int_0^{x}xf(x)dx, x \in \Omega$. The maximum point  $x_0$ can be obtained by $x_0=\text{arg}_x[\max \mathbb{E}
(x)]$.

(\uppercase\expandafter{\romannumeral2}) After we get the initial point $x_0$, we will generate $x_2,x_3...,x_n$ sequentially. Suppose we have $n$ points using ~\eqref{eqn:eqlabel10}. Then the $(n+1)$th point can be obtained by
\begin{align}
\label{eqn:eqlabel10q}
\begin{split}
 x_{n+1} & =\underset{x}{\text{arg}} \mathcal{E}_{\beta}(\Omega,N)  =\text{arg}\max_{\Omega} \min_{x_i,x_j}\left \{ \sum_{i=1}^{n-1}\sum_{j=i+1}^{n}\frac{\omega(x_i,x_j)}{\parallel x_i-x_j\parallel^m}  \right \}^{\frac{1}{m}}.
 \end{split}
\end{align}
(\uppercase\expandafter{\romannumeral3}) If $\left | x_{n+1}-x_n \right |\ge r_{\text{min}}(x_{1:n}^*)$, we further develop an acceptance criterion for $x_{n+1}$: Given $u\sim U(u\mid 0,1)$, if $\frac{\left | x_{n+1}-x_n \right |}{\left |x_n \right |}\geq u$, we accept $x_{n+1}$; otherwise, we reject it.

(\uppercase\expandafter{\romannumeral4}) After we get $n$ points, we can use some statistical techniques such as regression or kriging to estimate the underlying manifold, where the density can be updated with $\hat f (x)$, and $\gamma (x)=\hat \gamma(x)$, we can recursively continue to generate different configurations of discrete manifolds. 

\subsection{Pseudo-marginal Metropolis–Hastings Sampling}

Consider a hidden Markov model, described by $X_t \sim f_{\theta}(X_t \mid X_{t-1}),Y_t\mid X_t \sim g_{\theta}(y_t \mid X_t),$ given $x_0$, $X_t (t=1,2,...n)$ is a latent variable to be observed, the measurements ${Y_t}$ are assumed to be conditionally independent given ${X_t}$, the most objective is to estimate $\{X_{1:t}, \theta\}$. The Particle Metropolis-Hastings \cite{andrieu2010particle}, proposed an MCMC method to randomly "walk around" in the assumed measurable $\theta$ space and thus draw samples from the approximated posterior $\hat{p}(X_{1:t},\theta \mid y_{1:t})$, whose closed-form ${p}(X_{1:t},\theta \mid y_{1:t}) = {p}(\theta \mid y_{1:t}) \cdot {p}(X_{1:t} \mid y_{1:t}, \theta) $ is unreachable and cannot be evaluated pointwise exactly. 

We will introduce how Chebyshev particles are embedded for the following steps: For the parameter that locates at $\{\theta,X_{1:t}\}$, a new parameter $\{\theta',X_{1:t}^{'}\}$ is proposed from a proposal $q(\theta',X_{1:t}^{'}\mid \theta,X_{1:t})$ with the probability of acceptance
\begin{align}
\label{eqn:eqlabel1011q}
\begin{split}
%\alpha (\theta',\theta,X',X)& =\min\left \{ 1,\frac{p(X_{1:t}^{'},\theta'\mid y_{1:t})q(\theta,X_{1:t}\mid \theta',X_{1:t}^{'})}{p(X_{1:t},\theta\mid y_{1:t})q(\theta',X_{1:t}^{'}\mid\theta,X_{1:t} )} \right \}\\
\alpha & =\min\left \{ 1,\frac{p(X_{1:t}^{'},\theta'\mid y_{1:t})q(\theta,X_{1:t}\mid \theta',X_{1:t}^{'})}{p(X_{1:t},\theta\mid y_{1:t})q(\theta',X_{1:t}^{'}\mid\theta,X_{1:t} )} \right \}= \min\left \{ 1,\frac{p(y_{1:t}\mid \theta')p(\theta^{'})q(\theta\mid \theta')}{p(y_{1:t}\mid \theta)p(\theta)q(\theta'\mid \theta)} \right \}.
 \end{split}
\end{align}
The optimal importance density function that minimizes the variance of importance weights, conditioned upon $X_{t-1}^i$ and ${y_t}$ has been shown \cite{doucet2000sequential} to be 
\begin{align*}
\label{eqn:eqlabel1011aq}
\begin{split}
q(X_t\mid X_{t-1}^i,y_t)_{opt} & =p(X_t\mid X_{t-1}^i,y_t)=\frac{p(y_t\mid X_t,X_{t-1}^i)p(X_t\mid X_{t-1}^i)}{p(y_t\mid X_{t-1}^i)}.
 \end{split}
\end{align*}
While sampling from $p(y_t\mid X_t,X_{t-1}^i)$ may not be straightforward. As $X_{1:t}$ belongs to the "deterministic" part of the discrete manifolds of the space, $X_{1:t} \in \Omega$, %from Algorithms 1, 
the choice of importance density $ q(X_t|y_t,X_{t-1}^{a_{t-1}^{(i)}})$ is from the real configuration of the minimum energy, where $a_{t}^{(i)}$ denotes the ancestor of particle $X_t^i$%shown in Algorithm ~\ref{algo:event2}
. If $N\rightarrow \infty$, we have $\lim_{N\rightarrow \infty}q(X_t|y_t,X_{t-1}^{a_{t-1}^{(i)}})=p(X_t|X_{t-1}^i,y_t)$.
\iffalse
\begin{equation*}\label{eqn:eqlabel1193a}
\lim_{N\rightarrow \infty}q(X_t|y_t,X_{t-1}^{a_{t_1}^{(i)}})=p(X_t|X_{t-1}^i,y_t).
\end{equation*}
\fi
Thus, our proposal converges to the optimal importance density. We can obtain a stochastic estimator of $p(y_{1:T}\mid \theta)$. This likelihood can be estimated by the weights %from Algorithms 1 with
\begin{equation}\label{eqn:eqlabel119s3a}
\hat{p}_\theta(y_{1:T})=\prod_{t=1}^T(\frac{1}{N_x}\sum_{i=1}^{N_x}\frac{p(X_t|X_{t-1}^i,y_t)}{q(X_t|y_t,X_{t-1}^{a_{t-1}^{(i)}})}).
\end{equation}
It can be shown \cite{chopin2004central} that $\mathbb{E}[\hat{p}_\theta(y_{1:T})]=p_{\theta}(y_{1:T})$. The variance of the weights will be very small, this would be verified by the following experiments.

Combine ~\eqref{eqn:eqlabel1011q} and ~\eqref{eqn:eqlabel119s3a}, we can get the estimated acceptance ratio
\begin{equation*}\label{eqn:eqlaabel1319s3a}
%\hat{\alpha} (\theta',\theta,X',X)= \min\left \{ 1,\frac{\hat{p}(y_{1:t}\mid \theta')p(\theta^{'})q(\theta\mid \theta')}{\hat{p}(y_{1:t}\mid \theta)p(\theta)q(\theta'\mid \theta)} \right \}
\hat{\alpha}= \min\left \{ 1,\frac{\hat{p}(y_{1:t}\mid \theta')p(\theta^{'})q(\theta\mid \theta')}{\hat{p}(y_{1:t}\mid \theta)p(\theta)q(\theta'\mid \theta)} \right \}.
\end{equation*}
\iffalse
{ \color{black} 3.Variance minimum}

{ \color{black} 4.Experiments}
\fi
\section{Experiments}
%\section{Experiments}
%/home/daixiongming/Documents/Code/pmh-tutorial-2.1/r  

In this part, we will introduce the simulations where Chebyshev particles are embedded into the sequential Monte Carlo and its extension to Bayesian analysis for both the linear and non-linear models. We ran the experiments on an HP Z200 workstation with an Intel Core i5 and an $\#82-18.04.1-$ Ubuntu SMP kernel. The code is available at \url{https://github.com/986876245/ChebyshevParticles}.

\subsection{Linear Gaussian State Space Model}
%The quality of estimation using Sequential Monte Carlo can be used as an indicator for resampling quality. Here we discuss how to estimate the filtering distribution ${\pi_t(x_t)}_{t=0}^T$ for a linear Gaussian State-Space Model. The distributions of the filtered state composed of weighted particles can be used to estimate the likelihood.

The linear model is expressed by:
\iffalse
\begin{equation}
X_0\sim \mu(X_0),\ \ \
X_t\mid X_{t-1} \sim N(X_t;\phi X_{t-1},\delta _v^2),\ \ \
Y_t\mid X_t \sim N(y_t;X_t,\delta _e^2).
\end{equation}
\fi
\begin{equation*}\label{linearModel}
x_t\mid x_{t-1} \sim g(x_t|x_{t-1})dx_t,\ \ \
y_t\mid x_t \sim f(y_t|x_{t})dy_t+e_o.
\end{equation*}
Where $g(x_t|x_{t-1})=\phi x_{t-1}+e_v$, the noise from tracking $e_v \sim N(0,\delta _v^2)$, the noise from observations $e_o \sim N(0,\delta_o^2)$. Here we use ~\eqref{eqn:eqlabel10q}, to compute $ \widehat{x}^N_{0:T}$, and $\widehat{p}^N_\theta(y_{1:T})$ with Riesz particles instead for few evaluations. 
\begin{align*}
\label{eqn:eqlabel10g}
\begin{split}
 g(\widehat{x}_t|\widehat{x}_{t-1}) & =\underset{x}{\text{arg}} \mathcal{E}_{\beta}(\Omega,N')=\text{arg}\max_{x} \min_{x_i,x_j} \left \{ \sum_{x\neq \widehat{x}_i,i=1}^{t-1}\frac{\omega(\widehat{x}_i,x)}{\parallel \widehat{x}_i-x\parallel^m}  \right \}^{\frac{1}{m}},\\
  \omega (\widehat{x}_i,x)&\propto e^{\left [ \gamma (\widehat{x}_i) \gamma (x) +\beta \parallel \widehat{x}_i-x\parallel \right ]^{-\frac{m}{2d}}}.
 % & \omega (x_i,x_j)=\left [ \gamma (x_i) \gamma (x_j) +\beta \parallel x_i-x_j\parallel \right ]^{-\frac{m}{2d}}\omega' (x_i,x_j). \\
 %&  \omega' (x_i,x_j)=\Phi (\frac{\parallel x_i-x_j\parallel}{r_N}), \\
 %& r_N=C_NN^\frac{-1}{d}, \exists \beta, C_N\in R^+. \\
\end{split}
\end{align*}
For the linear Gaussian state space model, an optimal proposal distribution to propagate the particles $x^i_t, i=1,N$ can be derived \cite{dahlin2015getting} from
\begin{equation*}
\label{increasess}
\begin{split}
       p_{\theta}^{\text{opt}}(x_t^i \mid x_{t-1}^i,y_t) & \propto  g_{\theta}(y_t \mid x_t^i)f_{\theta}(x_t^i \mid x_{t-1}^i) =\mathbb{N}(x^i_t;\sigma^2[\sigma_o^{-2}y_t+\sigma^{-2}_v\phi x^i_{t-1}],\sigma^2)
\end{split}    
 \end{equation*}
with $\sigma^{-2}=\sigma_v^{-2}+\sigma_o^{-2}$. To ensure the stability of the algorithm and try to minimize the variance of the incremental particle weights at the current time step, we set $\gamma (x) \propto p_{\theta}^{\text{opt}}(x_t^i \mid x_{t-1}^i,y_t)$. The latent state $x_t$ can be estimated with an unbiased quantity $\widehat{x}^N_t=\frac{1}{N}\sum_{i=1}^{N}x^i_t$, here, $N$ is the number of particles for estimating the state, $N'$ is the number of Chebyshev particles for discretizing the submanifolds.  In order for the Chebyshev particles to be fully sampled, we give the indices with the remainder of $N$ divided by $N',(N>N')$ for the specific particles $x^i_t$.

\begin{figure*}[!htp]
	\centering
%	\begin{minipage}{.49\textwidth}
	%	\captionsetup{justification=centering,margin=0.5cm}
		\centering
		\includegraphics[width=.95\linewidth]{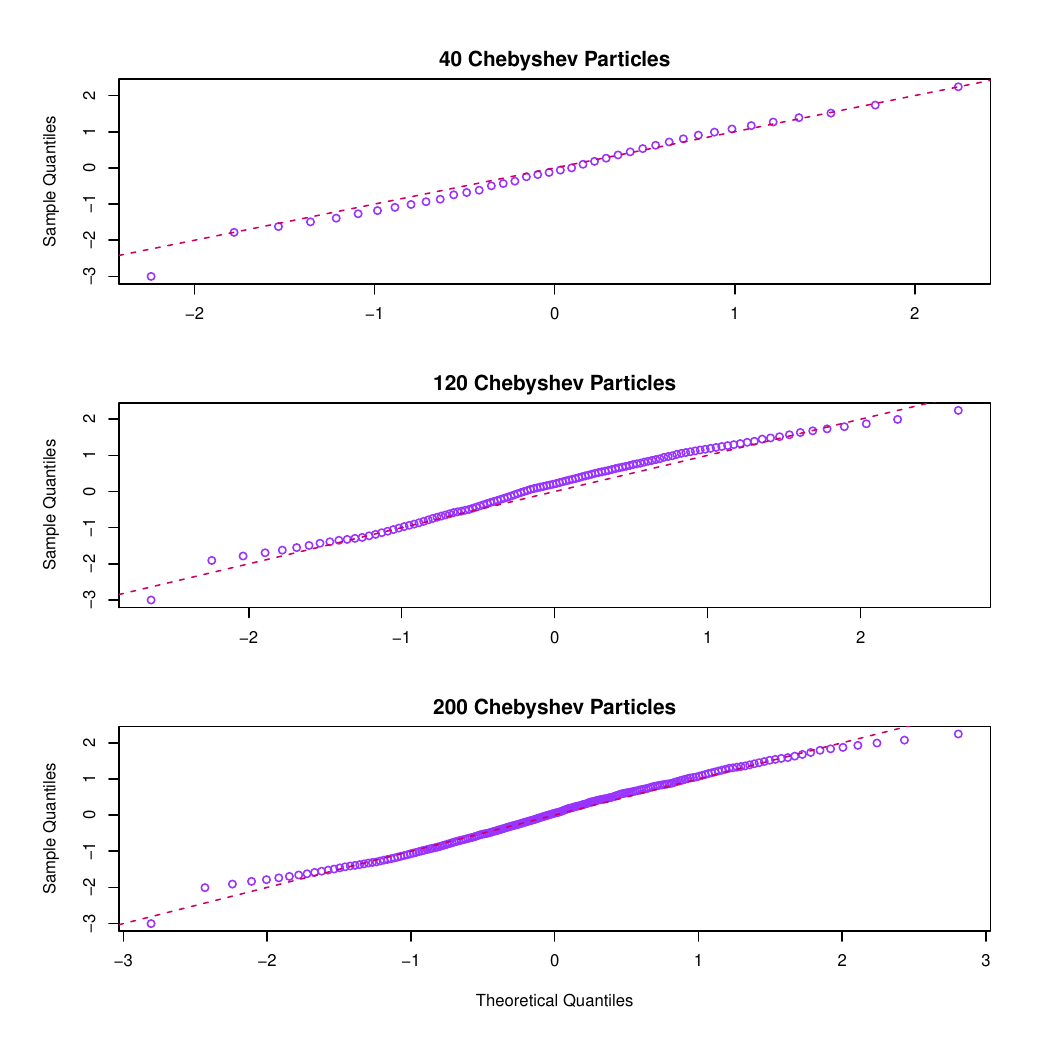}
		\caption{Theoretical Quantiles for 40, 120 and 200 Chebyshev particles.}
		\label{fig:experiment2a4g}
%	\end{minipage}
	
\end{figure*}

We first conduct experiments with different Chebyshev particles to discretize the submanifolds. These particles will approach different straight lines when they are mapped to a particular space shown in \autoref{fig:experiment2a4g} and satisfy uniform distribution from theorem 1, the parameters of the objective for Chebyshev particles: $\{\beta=1,m=40,d=1\}$.
%/home/daixiongming/Documents/Code/pmh-tutorial-2.1/r_chebyshev/qqnorm_total.r

Then, we embed these particles into sequential Monte Carlo, the states recursively interact in the Chebyshev particles set $\mathbb{P}$, to compare with the ground truth, we provide a simulated data record from the model with $T=250$ observations with initial value ${\phi=0.75,\delta_v=1.00,\delta_o=0.10}$ and $\widehat{x}_0=0$.  The estimated log-bias and log-MSE for the Chebyshev particles embedded in sequential Monte Carlo when varying the number of particles $N$ are %shown in \autoref{fig:experiment4a}. The corresponding table is 
shown in ~\autoref{tab:table_particles}.

\iffalse

\begin{figure}[!htp]
	\centering
	%\begin{minipage}{.49\textwidth}
		\captionsetup{justification=centering,margin=0.5cm}
		\centering
		\includegraphics[width=.95\linewidth]{example1_lgss_XMD_cut_c_finishe.png}
		\captionof{figure}{The difference between the Riesz particles embedded model with the ground truth.}
		\label{fig:experiment2a}
%	\end{minipage}
	
\end{figure}
\fi
\iffalse
%by run:/home/daixiongming/Documents/Code/pmh-tutorial-2.1/r/example1-lgss_XMD.R
\begin{figure*}[!htp]
	\centering
	%\begin{minipage}{.49\textwidth}
	%	\captionsetup{justification=centering,margin=0.5cm}
		\centering
		\includegraphics[width=.95\linewidth]{log_mse_bias_new.png}
		\caption{The estimated log-bias and log-MSE for varying Riesz particles embedded sequential Monte Carlo.}
		\label{fig:experiment4a}
%	\end{minipage}
	
\end{figure*}
\fi
\iffalse
\begin{figure}[!htp]
	\centering
	%\begin{minipage}{.49\textwidth}
		\captionsetup{justification=centering,margin=0.5cm}
		\centering
		\includegraphics[width=.95\linewidth]{example1_xMD_lgss_cut_a_fjinished.png}
		\captionof{figure}{Theoretical Quantiles for Riesz particles.}
		\label{fig:experiment4v}
%	\end{minipage}
	
\end{figure}
\fi
%/home/daixiongming/Documents/Code/pmh-tutorial-2.1/r/Rplot_LandsExample2_lgss_XMD.pdf  (is not for chebyshev)
%by run: /home/daixiongming/Documents/Code/pmh-tutorial-2.1/r_chebyshev/example2-lgss_XMD.R 
\begin{figure*}[!htp]
	\centering
	%\begin{minipage}{.49\textwidth}
	%	\captionsetup{justification=centering,margin=0.5cm}
		\centering
        \includegraphics[width=.95\linewidth]{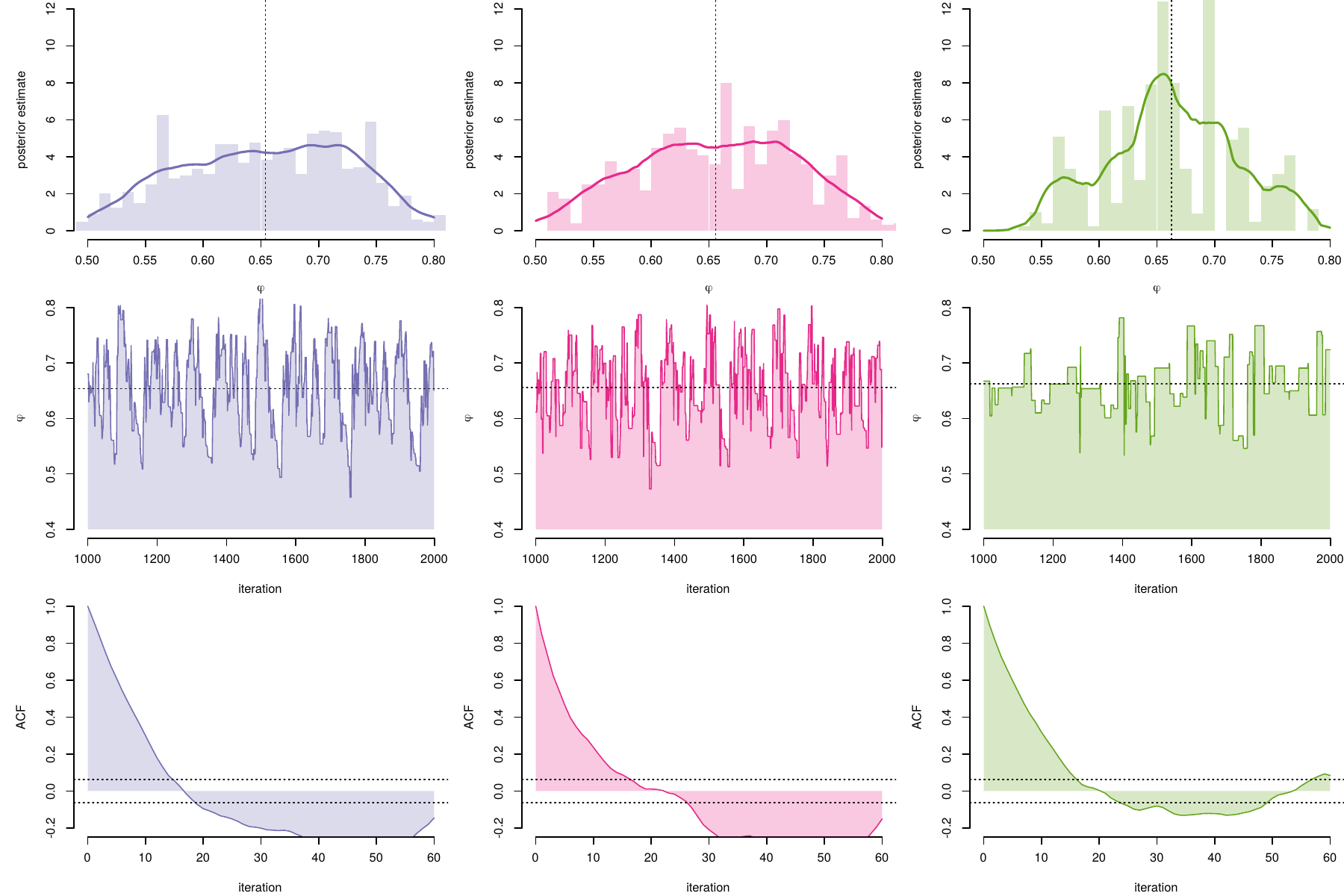}
		\caption{Posterior estimate, burning process and ACF for different step size:$h_1=0.05, h_2=0.1, h_3=0.5$.}
		\label{fig:experiment334}
%	\end{minipage}
	
\end{figure*}
%by run:/home/daixiongming/Documents/Code/pmh-tutorial-2.1/r_chebyshev/example1-lgss_XMD.R
\begin{table*}
\begin{center}
\begin{tabular}{|c|c|c|c|c|c|c|c|}
\hline
Number of particles(N) & 10 & 20 & 50 & 100 & 200 & 5000 & 1000 \\
\hline\hline
log-bias & -3.76 & -4.05 & -4.47 & -4.87 & -5.24 &-5.69 & -5.98 \\
log-MSE & -6.95 & -7.67 & -8.48 & -9.28 & -10.01 & -10.89 & -11.48 \\
\hline
\end{tabular}
\end{center}
\caption{The log-bias and the log-MSE of the filtered states under 200 Chebyshev particles for varying N.}
\label{tab:table_particles}
\end{table*}
Here, we extend Chebyshev particles into the pseudo-marginal Metropolis-Hastings algorithm provided in Section 3.2 for the Bayesian parameter inference of hidden Markov models. We estimate the posterior for $\phi$, $\phi \in (-1,1)$ describes the persistence of the state, and keep ${\delta_v=1.00, \delta_e=0.10}$ fixed, the prior for $\phi_0=0.75$, and specify the number of Chebyshev particles in the set as $100$, which is far less than the iterations ($\geq 2000$), from this point, we have largely scaled the particle sets for our model, then, we just need few evaluations to infer this model. We conduct different step sizes, $h_1=0.05,h_2=0.1,h_3=0.5$, the posterior estimate, the burning process and plots of the auto-correlation of a time series by lag are shown in \autoref{fig:experiment334}. The corresponding table is shown in \autoref{tab:table_particles2}.

%/home/daixiongming/Documents/Code/pmh-tutorial-2.1/r_chebyshev/extra-code-for-tutorial/example2-lgss-varyingT_XMD.R
\begin{table*}
\begin{center}
\begin{tabular}{|c|c|c|c|c|c|c|}
\hline
Number of particles(N) & 10 & 20 & 50 & 100 & 200 & 500 \\
\hline\hline
Estimated posterior mean & 0.559 & 0.769 & 0.737 & 0.696 & 0.709 & 0.717 \\
Estimated posterior variance & 0.105 & 0.039 & 0.023 & 0.012 & 0.005 & 0.001 \\
\hline
\end{tabular}
\end{center}
\caption{The estimated posterior mean and variance when varying T.}
\label{tab:table_particles2}
\end{table*}

\subsection{Nonlinear State Space Model}
We continue with a real application of our proposal to track the stochastic volatility, a nonlinear State Space Model with Gaussian noise, where log volatility considered as the latent variable is an essential element in the analysis of financial risk management. %. 
The stochastic volatility is given by
\begin{align*}
\label{eqn:eqlabel10gaa}
\begin{split}
&X_0\sim N(\mu,\frac{\sigma _v^2}{1-\rho  ^2}),\ \ X_t\mid X_{t-1} \sim N(\mu+\rho(X_{t-1}-\mu),\sigma _v^2),\ \ Y_t\mid X_t \sim N(0,exp(X_t)\tau),
\end{split}
\end{align*}
\iffalse
\begin{equation}
\begin{aligned}
  & X_0\sim N(X_0;\mu,\frac{\sigma _v^2}{1-\rho  ^2})\\
  & X_t\mid X_{t-1} \sim N(X_t;\mu+\rho  (X_{t-1}-\mu),\sigma _v^2)\\
  & Y_t\mid X_t \sim N(y_t;0,exp(X_t)\tau) 
\end{aligned}
\end{equation}

\fi
where the parameters $\theta =\left \{\mu, \rho  ,\sigma_v,\tau \right \}$, $\mu\in \mathbb{R},
\rho  \in [-1,1]$, $\sigma _v $ and $\tau \in \mathbb{R}_+$, denote the mean value,
the persistence in volatility, the standard deviation of the state process and the instantaneous
volatility, respectively. 

The observations $y_t=\log(p_t/p_{t-1})$,  also called log-returns, denote the logarithm of the daily difference in the exchange rate $p_t$, 
%more specific,
%\begin{equation}
%y_t=log[\frac{p_t}{p_{t-1}}]=100[log(p_t)-log(p_{t-1})]
%\end{equation}
%where $s_t$ denotes the price of some financial asset at time $t$. 
here, $\{{p_t}\}_{t=1}^T$ is the daily closing prices of the NASDAQ OMXS30 index (a weighted average of the 30 most traded stocks at the Stockholm stock exchange) \cite{dahlin2015getting}. We extract the data from {\color{blue} \href{https://www.quandl.com/}{Quandl}} for the period between January 2, 2015 and January 2, 2016. The resulting log-returns are shown in \autoref{fig:experimentu4aa}. We use SMC to track the time-series persistency volatility,  large variations are frequent, which is well-known as volatility clustering in finance, from the equation (42), as $|\phi|$ is close to $1$ and the standard variance is small, the volatility clustering effect easier occurs. Here, the parameters of the objective for Chebyshev particles: $\{\beta=1,m=40,d=1\}$, the size of Chebyshev particles is $200$. The initial value is $\mu_0=0, \sigma_0=1, \phi_0=0.95, \sigma_{\phi}=0.05, \delta_{v_0}=0.2, \sigma_{v}=0.03$ % { \color{black}We keep the same parameters }as \cite{dahlin2015getting},
%where $\mu \sim N(0,1), \phi \sim TN_{[-1,1]}(0.95,0.05^2)$, $ \delta _v \sim \hbox{\it Gamma}(2,10) $, $\tau=1$.
We obtain good performance that the posterior estimation can be inferred from a few evaluations, it greatly scales the computation load for high-dimensional sampling, shown in \autoref{fig:experimentu4aa}.
\iffalse
%/home/daixiongming/Documents/Code/Minimal_Energy_Way/1323273/MED Rcodes/multinorm_XMD.R
\begin{figure*}[!htp]
	\centering
	%\begin{minipage}{.49\textwidth}
	%	\captionsetup{justfification=centering,margin=0.5cm}
		\centering
		\includegraphics[width=.95\linewidth]{example3-sv_XMD_landscape.pdf}
		\caption{Top: The daily log-returns and estimated log-volatility with $95\%$ confidence intervals of the NASDAQ OMXS30 index for the period between February 4, 2015 and February 4, 2016. Bottom: the posterior estimate(left), the trace of the Markov chain(middle) and the corresponding ACF(right) of $\mu$(purple), $\phi$(magenta) and $\sigma_v$(green) obtained from Chebyshev particles embedded PMH. The dotted and solid gray lines in the left and middle plots indicate the parameter posterior mean and the parameter priors, respectively.}
		\label{fig:experimentu4aa}
%	\end{minipage}	
\end{figure*}
\fi
%%/home/daixiongming/Documents/Code/pmh-tutorial-2.1/r/normal distribution_forExample3.R
%/home/daixiongming/Documents/Code/pmh-tutorial-2.1/r/example3-sv_XMD_phd_recover_new.R
%/home/daixiongming/Documents/Code/pmh-tutorial-2.1/r/figures/example3-sv_XMD.pdf
\begin{figure*}[!htp]
	\centering
	%\begin{minipage}{.49\textwidth}
	%	\captionsetup{justfification=centering,margin=0.5cm}
		\centering
		\includegraphics[width=.95\linewidth]{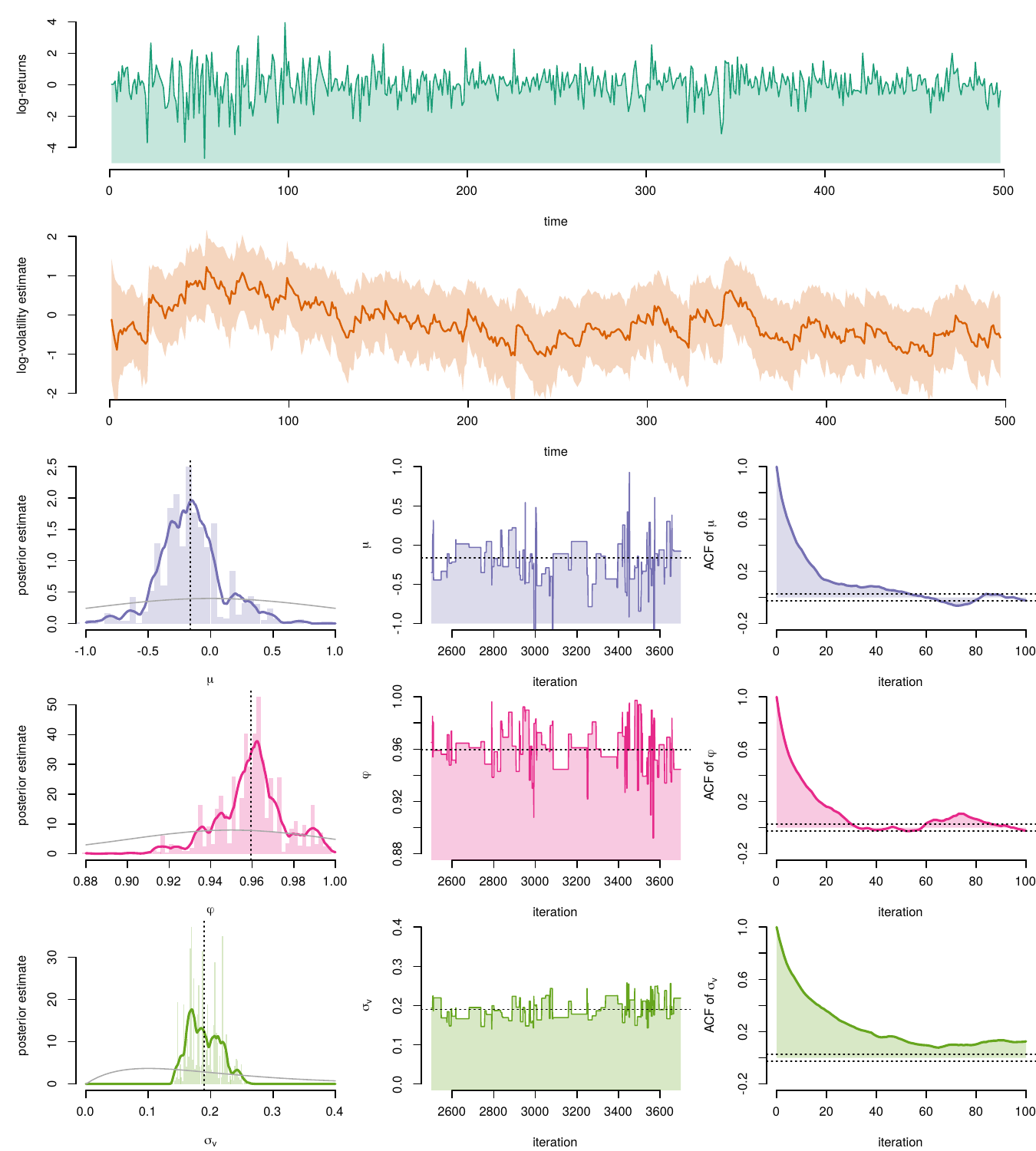}
		\caption{Top: The daily log-returns and estimated log-volatility with $95\%$ confidence intervals of the NASDAQ OMXS30 index for the period between February 4, 2015 and February 4, 2016. Bottom: the posterior estimate(left), the trace of the Markov chain(middle) and the corresponding ACF(right) of $\mu$(purple), $\phi$(magenta) and $\sigma_v$(green) obtained from Chebyshev particles embedded PMH. The dotted and solid gray lines in the left and middle plots indicate the parameter posterior mean and the parameter priors, respectively.}
		\label{fig:experimentu4aa}
%	\end{minipage}
	
\end{figure*}
\section{Conclusion}
Markov chain Monte Carlo (MCMC) provides a feasible method for inferring Hidden Markov models.  However, it is often computationally prohibitive and especially constrained by the curse of dimensionality, since the Monte Carlo sampler traverses randomly taking small steps within uncertain regions in the parameter space. In this process, a large number of duplicate samples will be burned, and these duplicate samples greatly increase the computational load. We have introduced a deterministic sampling mechanism, in which all generated samples are derived from particle interactions under a weighted Riesz polarization maximizing criterion. All samples inherit the properties of both a well-separated distance and a bounded covering radius. We have embedded them into MCMC, where we have achieved high performance in our experiment of a hidden Markov model. Only a few evaluations are required, and we can extend our method into high-dimensional sampling. %Deterministic sampling is a popular mechanism to find an optimally determined region of interest in the design of computer experiments, where deterministic points distributed geometrically reflect the special representations of the region. 
%We consider the posterior distribution of the objective as a mapping of samples in an infinite-dimensional Euclidean space where deterministic submanifolds are embedded,  and propose a new flexible minimum energy criterion, termed as weighted Riesz $\beta$-energy, to discretize rectifiable submanifolds of interest.  We study the characteristics of weighted Riesz particles and embed them into the MCMC, a new sampler with a high acceptance ratio, and a few evaluations are proposed. We achieve good performance from the experiments of a hidden Markov model in both the linear and non-linear cases. 
For future research, we will develop a kernel for the Chebyshev particles and scale the model with low complexity of computations from the perspective of equilibrium states on high-dimensional sampling.

\section*{Acknowledgments}
This was supported in part by BRBytes project.

%Bibliography
\bibliographystyle{unsrt}  
\bibliography{references}

\end{document}